\documentclass[11pt]{article}
\usepackage{etex} 

\usepackage[final]{acl}
\usepackage{booktabs}
\usepackage{times}
\usepackage{tabularx}
\usepackage{latexsym}
\usepackage{amsmath}
\usepackage{amsfonts}
\usepackage{multirow}
\usepackage{longtable}
\usepackage{lscape}
\usepackage{arydshln}
\usepackage[T1]{fontenc}

\usepackage[utf8]{inputenc}

\usepackage{microtype}

\usepackage{inconsolata}

\usepackage{graphicx}
\usepackage{float}
\newcommand{\tbf}[1]{\textbf{#1}}
\newcommand{\ul}[1]{\underline{#1}}
\title{MAPLE: Enhancing Review Generation with Multi-Aspect Prompt LEarning in Explainable Recommendation}

\author{
  Ching-Wen Yang\thanks{P76114511@gs.ncku.edu.tw},
  Zhi-Quan Feng\thanks{P78123011@gs.ncku.edu.tw},
  Ying-Jia Lin\thanks{yjlin@cgu.edu.tw},
  Che-Wei Chen\thanks{Q56104076@gs.ncku.edu.tw},\\
  \textbf{Kun-Da Wu} \thanks{harrisonwu@google.com},
  \textbf{Hao Xu}\thanks{hao.xu@nyu.edu},
  \textbf{Jui-Feng Yao}\thanks{eagley@google.com},
  \textbf{Hung-Yu Kao}\thanks{hykao@cs.nthu.edu.tw} \\
  Department of Computer Science and Information Engineering, \\
  National Cheng-Kung University, Google Pixel Software
}
\begin{document}

\maketitle

\begin{abstract}
  The Explainable Recommendation task is designed to receive a pair of \textit{user} and \textit{item} and output explanations to justify why an item is recommended to a user.
  Many models approach review generation as a proxy for explainable recommendations. While these models can produce fluent and grammatically correct sentences, they often lack precision and fail to provide personalized, informative recommendations.
  To address this issue, we propose a personalized, aspect-controlled model called Multi-Aspect Prompt LEarner (MAPLE), which integrates aspect category as another input dimension to facilitate memorizing fine-grained aspect terms. Experiments conducted on two real-world review datasets in the restaurant domain demonstrate that MAPLE significantly outperforms baseline review-generation models. MAPLE excels in both text and feature diversity, ensuring that the generated content covers a wide range of aspects. Additionally, MAPLE delivers good generation quality while maintaining strong coherence and factual relevance.
  The code and dataset used in this paper can be found here \footnote{https://github.com/Nana2929/MAPLE.git}.
\end{abstract}

\section{Introduction}
In the context of Natural Language Generation (NLG) explainable recommendation models, a good explanation is required to have the following characteristics: 1) \textit{Diversity}: for the same item, a model should generate personalized rationales for different users. 2) \textit{Factuality}: the recommended feature or content should be factually relevant to the item. 3) \textit{Precision}: the recommended feature should be as precise (as opposed to general) as possible.

\begin{figure}[t]
  \centering
  \includegraphics[width=0.45\textwidth]{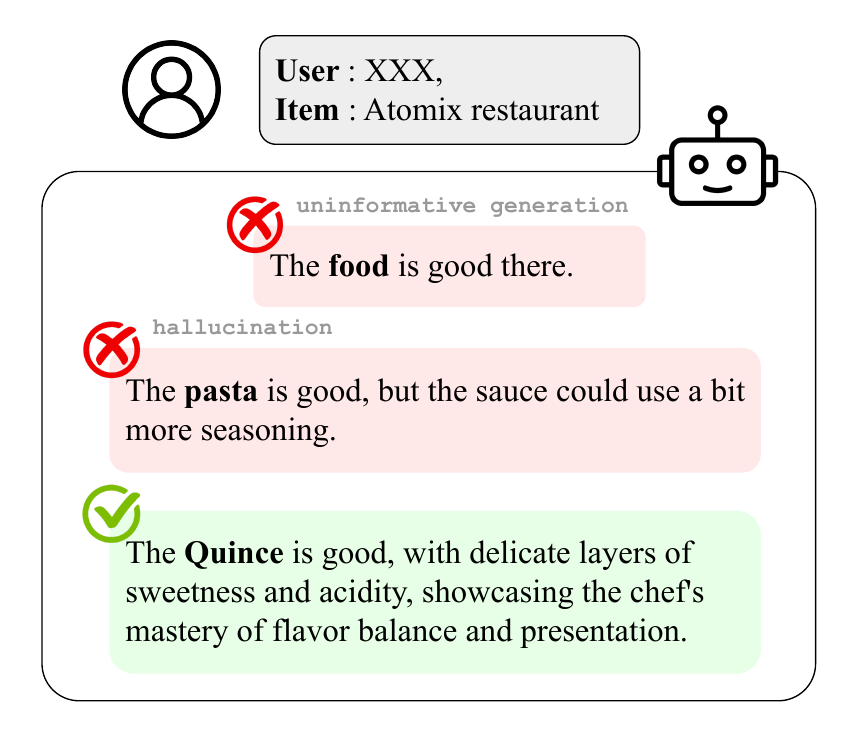}
  \caption{Overview of explainable recommendation: A simple illustration of the "diversity-factuality dilemma".}
  \label{fig:introduction}
\end{figure}

Within these criteria, we observe a "diversity-factuality dilemma", as illustrated in Figure \ref{fig:introduction}. Models that generate generic aspects, such as "food" in the context of restaurants, may provide safe and broadly applicable explanations but risk being uninformative and repetitive, which has been noted in prior studies \citep{Li2017nrt, Li2017att2seq, li2021peter, li2023pepler}. On the other hand, models creating diverse and creative aspects produce more engaging explanations but may compromise factual relevance by including irrelevant details, which is referred to as the hallucination problem. Few existing models effectively balance these traits.

To address this, we introduce the concept of "multi-aspect" from multi-aspect sentiment analysis \citep{Lu2011MASA, xianghua2013multi}, which assumes a limited inventory of multiple, distinct aspects like “food quality” or “service” in the restaurant domain, providing a more fine-grained understanding of user preferences. In this paper, we propose a simple yet effective two-stage tuning approach that integrates aspect as an auxiliary signal to improve the memorization of rich aspect terms, called MAPLE (Multi-Aspect Prompt LEarner). For \textit{diversity}, we utilize a Multilayer Perceptron (MLP) model to predict aspects and employ a distribution-balanced loss function \citep{Wu2020dbloss} to ensure diverse predictions. For \textit{factuality}, we generate recommendation text using a retriever-reader framework, and for \textit{precision}, we retrieve specific information to guide the LLM in generating user- and item-specific recommendation text. Our contributions are as follows:
\begin{enumerate}
  \item MAPLE increases diversity, factuality, and precision of generated features, endorsed by the self-crafted explainability metrics from aspect-wise perspectives.
  \item MAPLE's generated explanations serve as good queries within the retriever-reader framework. By comparing with a latent personalized retriever model, we show that MAPLE more accurately predicts aspect relations, as evidenced by the results and case studies.
  \item We renew the review-generation datasets in the restaurant domain \citep{li2020nete} to include higher quality aspect terms and additionally label associated aspect categories, enhancing the research in this field.
\end{enumerate}

We treat MAPLE as a retriever in the retriever-reader framework with an LLM as the reader in Appendix \ref{appendix:case-study}, demonstrating that MAPLE's explanations combined with the LLM's comprehension yield enriched, personalized results.

\section{Related Works}
\subsection{Review Generation}
Past works \citep{Li2017att2seq, Li2017nrt, li2021peter, li2023pepler} have proposed end-to-end frameworks for learning a short explanation (often part of the review) for a user-item pair. Experimental statistics and case studies show that these models generate repetitive, overly generic sentences that are still far from good rationales \citep{xie2023prag}, and even suffer from hallucination issue \citep{xie2023prag, maynez2020faithfulness} \footnote{A model exhibiting hallucinations generates content that includes inaccuracies or irrelevant information.}. It is also observed that the uninformative sentences are often due to models' being incapable of generating precise and informative aspect terms.

\subsection{Aspect-aware Explanation Generation}
According to \citet{Zhang2023ABSASurvey} which studies into aspect-based sentiment analysis, aspect categories \(c\) refer to broad attributes (such as \textit{food} or \textit{service} in the restaurant domain) and aspect terms \(a\) refer to specific targets (such as \textit{the beef tacos} in the review "The beef tacos here are amazing").

To address overly general aspect terms, \citet{Ni2018ExpansionNet} proposed ExpansionNet, which expands short phrases into detailed explanations via a tri-encoder framework that predicts aspect-term importance using an aspect encoder. Aspect terms are obtained through ABAE \citep{He2017ABAE}, an unsupervised model that infers aspect categories and retrieves top-K aspect terms.

Building on ExpansionNet, UARM \citep{Sun2022UARM} enhances the use of inferred aspect embeddings from ABAE by converting them into aspect-aware user and item representations.
\citet{li2020nete} extracts review aspect terms with Sentires \citep{zhang2014sentires}, and its proposed model NETE does aspect-term condition generation in training. In the inference stage, NETE uses Point-wise Mutual Information (PMI) to rank and select an aspect term by user preference. However, aspect-term distribution might be too sparse to be captured by PMI.  Therefore, there's an emerging line of works that simply assume the ground-truth aspect terms are given: One of them is PETER+ \citep{li2021peter}, which places aspect-term tokens behind user and item IDs to guide the generation of explanations. Another example is ERRA \citep{ERRA2023explainable}, which leverages Sentence-BERT \citep{SentenceBERT} to encode aspect terms and the corpus. It searches for appropriate aspect and corpus vectors to support the text generation. 

\begin{figure*}[t]
  \centering
  \includegraphics[width=0.8\textwidth]{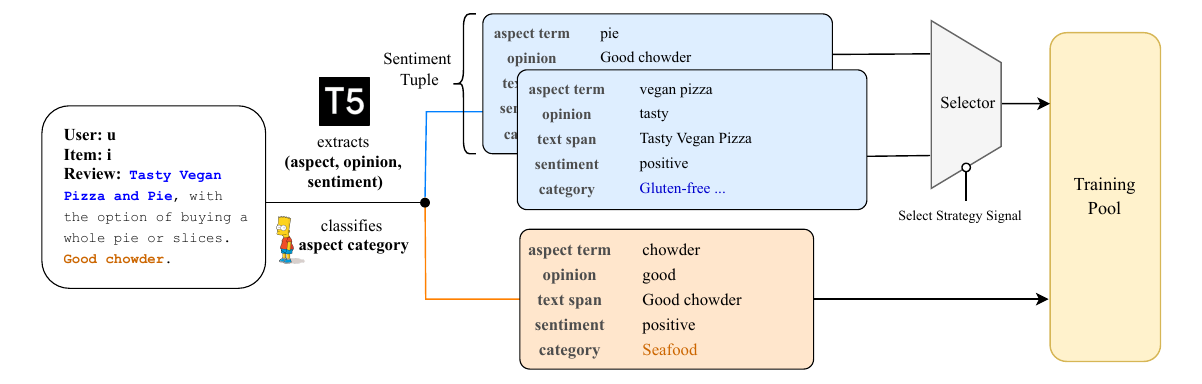}
  \caption{Automated sentiment-analysis pipeline for Multi-Aspect Review Segmentation.}
  \label{fig:sentiment-pipeline}
\end{figure*}

Our approach aligns more with the line of NETE, wherein in the training stage, MAPLE is trained by a given \textit{aspect-category} signal, and in the inference stage, it decides which aspect signal it wants to use. Compared to ERRA, MAPLE fine-tunes the generation model and incorporates multi-aspect learning to predict aspects with greater diversity. This makes MAPLE a simple yet effective approach for controlling from a broad sense and generating fine-grained aspect terms.

\subsection{Retrieval-Augmented Explanation Generation}

To address the issue of factual relevance, a long-existing challenge in NLG task, \citet{xie2023prag} proposes a retriever-reader two-stage framework called PRAG. In the first stage, a personalized retriever formulates a latent query based on the input user and item; in the second stage, a reader model generates the explanation, grounding it in the retrieved content, which consists of past reviews that are inherently factual. Given the current advancements in LLM \cite{10598017}, we argue that LLMs fit into the role of the reader very well as their reading comprehension ability can be leveraged not only to enhance factual relevance but also to refine the style of the generated reviews, making them more persuasive and closely aligned with real explanations.
In our case studies in Appendix \ref{appendix:case-study}, with the examples of employing MAPLE as a discrete retriever and LLM as a reader, we showcase the potential for improving both factual accuracy and the persuasiveness of explanations.

\section{Methodology}
In this work, we propose a two-stage training approach. We initially extract the aspect term from reviews in Section \ref{sec:Multi-aspect Review Segmentation} to support the two-stage training. In Stage 1 (Section \ref{sec:maple_stage1}), we focus exclusively on the explanation generation task, optimizing it using the negative log-likelihood loss \( \mathcal{L}_T \) until convergence. In Stage 2 (Section \ref{sec:maple_stage2}), we introduce the recommendation loss, weighted by $\alpha$, into the total loss to equip the ID embeddings with the selector ability. Finally, our inference process is presented in Section \ref{subsec:inference}.

\subsection{Problem Setup}
\label{sec:problem-setup}
Consider a set of users \( U \) and items \( I \). For each user \( u \in U \) and item \( i \in I \) in the training set, there is an associated review $r$ covering several aspects $c_i$, each explained in terms of that aspect. The task of joint aspect-recommendation-explanation is to learn a function \(rec : (u, i) \rightarrow (\hat{c}_{u, i}, \hat{E}_{u, i})\), where \(\hat{c}_{u, i}\) is the predicted aspect-category distribution $\in \mathcal{R}^{n_\text{aspect}}$ ($n_{\text{aspect}}$ denotes the number of aspect categories) and \(\hat{E}_{u,i}\) is the textual explanation given to justify the recommendation of item $i$ to user $u$. In this context, \(\hat{c}_{u,i}\) serves as a by-product that is used in assisting the generation of \( \hat{E_{u,i}}\) in the inference stage.

\begin{figure*}[t]
  \includegraphics[width=\textwidth]{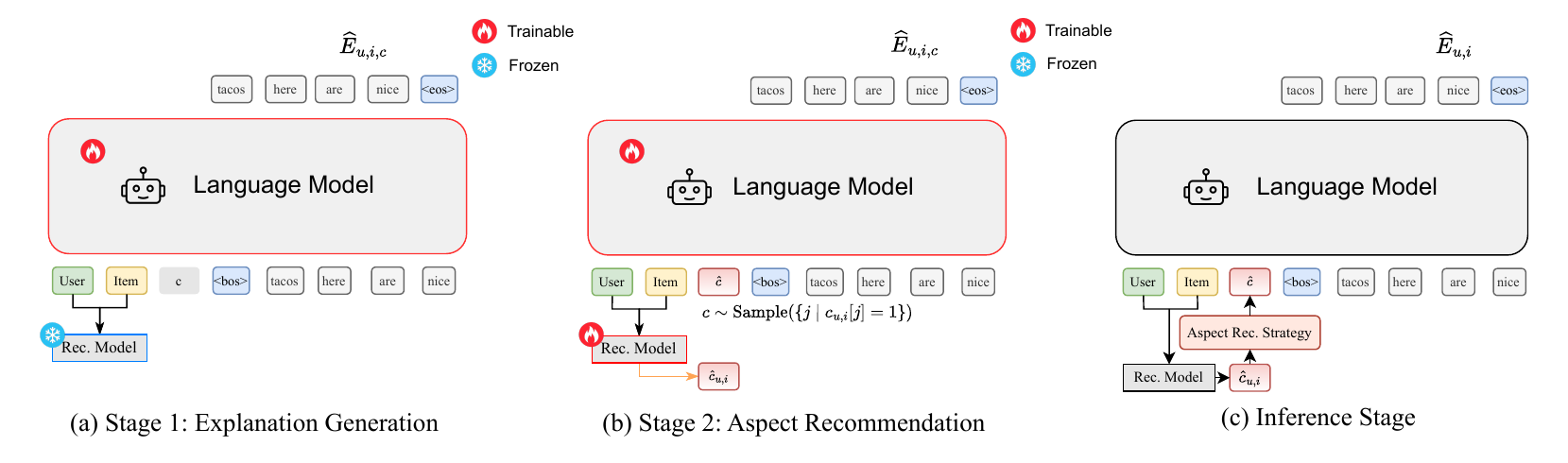}
  \centering
  \caption{The proposed MAPLE architecture. MAPLE is trained on Stage 1: Explanation Generation and Stage 2: Aspect Recommendation as a unified architecture. During inference, MAPLE exploits the trained Aspect-Recommendation Model (Rec. Model) in Stage 2 to predict an aspect distribution $\hat{c}_{u, i}$ and obtain an aspect signal $\hat{c}$, and then feed it back to Stage 1 for generation.}
  \label{fig:maple-architecture}
\end{figure*}
\subsection{Multi-Aspect Review Segmentation}
\label{sec:Multi-aspect Review Segmentation}
Instead of predicting aspect-term importance, as done by ExpansionNet and UARM, our model takes a broader perspective by learning aspect topics and extracting aspect terms from within these topics based on ID information. This approach prevents the model from overfitting to the characteristics of individual users or items, avoiding overly generic explanations that are conditioned on ID alone. By introducing aspect prompts $c$, the model focuses on learning the fine-grained aspect terms of the particular category, while the ID embeddings are learned as \textit{selectors} for relevant aspects.

We employ an automated sentiment-analysis pipeline to extract sentiment tuples \textit{(aspect term, opinion, sentiment, category)} from reviews. Concretely, we use a fine-tuned sentiment analysis model \citep{raffel2020T5} to extract sentiment tuples within the reviews and discard the reviews without tuples. We then use a zero-shot classifier \citep{Lewis2020bart} to assign each extracted aspect term with an aspect category (for details on model and aspect inventory, see Appendix \ref{appendix:dataset-models} and \ref{appendix:dataset-aspectlist}).
Finally, for every user-item pair, we check if there are multiple mined aspect terms under the same aspect category; if yes, we choose the tuple with an arbitrary select strategy, e.g., choose the one that is the longest, and add the associated tuple into the training data. Figure \ref{fig:sentiment-pipeline} illustrates an example of a review and its extracted sentiment tuples, Since we take one tuple for each category, this review yields two tuples. For the "Gluten-Free, Vegan, Vegetarian" category, we take ("vegan pizza", ...) since it has a longer aspect term compared to "pie"; for the "Seafood" category, we take ("chowder", ...).

\subsection{Stage 1: Explanation Generation}
\label{sec:maple_stage1}
We adopt the continuous prompt learning approach proposed by \citet{li2023pepler}. The input sequence can be represented as $S=[u, i, c, \texttt{<bos>}
  , e_1, \dots, e_{|E_{u,i,c}|}, \texttt{<eos>}
  ]$, where \texttt{<bos>} and \texttt{<eos>} are arbitrary tokens marking the beginning and end of a sentence, as shown in Figure \ref{fig:maple-architecture} (a).  Specifically, we prepare three sets of token embeddings $U\in \mathbb{R}^{|\mathcal{U}| \times d}; I\in \mathbb{R}^{|\mathcal{I}| \times d}; C\in \mathbb{R}^{|\mathcal{C}| \times d}$, where $\mathcal{U}, \mathcal{I}, \mathcal{C}$ represent the set of users, items and the predefined aspect categories of a dataset, respectively. We illustrate the details using aspect category because it is the same for obtaining ID representations. To obtain the representation of aspect category $c$, denoted $\mathbf{c}$, we index the token embeddings using the one-hot vector $g(c) \in \{0,1\}^{|\mathcal{C}|}$.

We then train MAPLE to condition the user-item-aspect signals for text generation.
We minimize the negative log-likelihood loss $L_T$, where $\mathcal{T}$ denotes the training set; $E_{u, i, c}$ denotes the explanation segment for the user-item-aspect pair; $c_{3+t}^{e_t}$ is offset by 3 to accommodate the user, item, and aspect prompt tokens. As discussed in Section \ref{sec:Multi-aspect Review Segmentation}, rather than learning just one explanation for a user-item pair, a user-item can now pair with different aspect categories $c_i$ depending on the categories labeled from the ground-truth review.
\begin{equation}
  \mathcal{L}_T = \frac{1}{|\mathcal{T}|} \sum_{(u,i,c) \in \mathcal{T}} \frac{1}{|E_{u,i,c}|} \sum_{t=1}^{|E_{u,i,c}|} -\log c_{3+t}^{e_t}
  \label{eq:L_T}
\end{equation}
\subsection{Stage 2: Aspect Recommendation}
\label{sec:maple_stage2}
During training, MAPLE takes a user-item pair and the auxiliary aspect category. However, in the inference stage, the aspect information may be absent. To close the gap between training and inference, we design an auxiliary task, called aspect recommendation, to recover the relationship between IDs and their associated aspect categories. The input of the aspect-recommendation task is a pair of user, item IDs, the output is the predicted aspect category probabilities $\hat{c_{u, i}} \in \mathbb{R}^{n_{\text{aspect}}}$. We implement the model with an MLP architecture:

\begin{equation}
  \label{eq:mlp}
  \hat{c}_{u,i} = \sigma\left(\mathbf{W}_c \delta(\mathbf{W}_h [\mathbf{u}; \mathbf{i}] + \mathbf{b}_h) + \mathbf{b}_c\right)
\end{equation}

In equation \ref{eq:mlp}, \(\mathbf{W}_h \in \mathbb{R}^{h \times 2d}\) and \(\mathbf{b}_h \in \mathbb{R}^{h}\) are the weights and biases of fully-connected layer(s), \, \(\mathbf{W}_c\in \mathbb{R}^{n_{\text{aspect}} \times h}\) and \(\mathbf{b}_c \in \mathbb{R}^{n_{\text{aspect}}}\) are the weights and biases of the output classifier layer, and $\delta$ and $\sigma$ are the activation functions ReLU and Sigmoid, respectively.

However, the aspect-category label exhibits a very skewed, long-tail distribution (Figure \ref{fig:long-tail}), which poses a challenge in training the aspect-recommendation model, since a conventional classifier easily overfits the head classes. We employ the distribution-balanced loss function \citep{Wu2020dbloss} to address this issue, effectively capturing aspect preferences from ID information rather than label distribution. It is formalized as:

\begin{align} \label{eq:Delta_i}
   & \mathcal{L}_{DB}(x^k, y^k) = \frac{1}{n_{\text{aspect}}}\sum_{i=0}^{n_{\text{aspect}}} \hat{r}^k_i \left[ y^k_i \log\left(1 + e^{-\Delta_i}\right) \right. \notag \\
   & \quad \left. + \frac{1}{\lambda} (1 - y^k_i) \log\left(1 + e^{\lambda \Delta_i}\right) \right]
\end{align}
\vspace{-1em} 
\begin{equation} \notag
  \Delta_i = z^k_i - \nu_i
\end{equation}

In equation \ref{eq:Delta_i}, \(\mathcal{L}_{DB}(x^k, y^k)\) represents the distribution-balanced loss function for the \(k\)-th sample; \(\hat{r}^k_i\) represents the sampling weighting factor for the \(i\)-th aspect class of the \(k\)-th sample; \(y^k_i\) is the ground truth label for the \(i\)-th class of the \(k\)-th sample; \(z^k_i\) is the logit for the \(i\)-th class of the \(k\)-th sample; \(\nu_i\) is the class bias for \(i\)-th class, and \(\lambda\) is a hyperparameter regularizing sigmoid's gradient issue.


In this stage, we add the recommendation loss weighted by $\alpha$ to the total loss, giving ID embeddings selector ability. Formally,
\begin{equation}
  L = L_T + \alpha L_{DB}
  \notag
\end{equation}
We conducted preliminary tuning and found that $\lambda=1$ yielded stable training, while larger values (e.g., $\lambda=5$) led to underfitting. Thus, we use $\lambda=1$ in all the reported experiments.

\subsection{Inference Stage}
\label{subsec:inference}

Given user and item IDs, we first infer stage 2's aspect-recommendation model and trim the predicted aspect distribution $\hat{c}_{u, i}$ to leave only the top 5, and then we sample $K (K \leq 5) $ aspects with replacement from the trimmed distribution, using the predicted probabilities as sample weights. The $K$ embeddings are then fused (by taking their mean) to form the aspect signal at position $\hat{c}$, as shown in Figure \ref{fig:maple-architecture} (b). The trained transformer is thus conditioned on $u, i,\hat{c}$ to generate a well-rounded explanation $\hat{E}_{u, i}$.

\section{Experiment Setup}
\subsection{Dataset}
In our experiments, we focus on the restaurant domain by benchmarking against two datasets from Yelp, which we refer to as \tbf{Yelp19} \cite{li2020nete} and \tbf{Yelp23} \citep{yelp2023dataset}, distinguishing them based on the year the data was provided. Different restaurants offer a wide variety of unique items (e.g., specific dishes like \textit{tonkatsu ramen} for a Japanese ramen store; \textit{risotto} for an Italian restaurant). This distinctiveness in aspect terms allows for more precise and compelling explanations for a specific restaurant. We provide the dataset sources, statistics, and details of how we renew the datasets using the automated sentiment pipeline in Appendix \ref{appendix:dataset-prep}.

\begin{table*}[t]
  \centering
  \renewcommand\arraystretch{1.05}
  \resizebox{0.85\textwidth}{!}{%
    \begin{tabular}{c|c|ccc|cccccc|c}
      \hline
      \multicolumn{12}{c}{\tbf{Yelp19}}                                                                                                                                                                                                                                            \\ \hline
      \multicolumn{1}{c|}{} & \multicolumn{1}{c|}{Factuality} & \multicolumn{3}{c|}{Aspect-wise Exp.} & \multicolumn{6}{c|}{Text Diversity} & \multicolumn{1}{c}{Gen. Quality}                                                                                                     \\ 
      \tbf{Method}          & \tbf{iFMR}                      & \tbf{FCR}                             & \tbf{iFCR}                          & \tbf{GT-FMR}                     & \tbf{USR}   & \tbf{uUSR}  & \tbf{iUSR}  & \tbf{D-2}   & \tbf{D-3}   & \tbf{ENTR}   & \tbf{MAUVE}  \\ \hline
      Att2seq               & 0.693                           & 0.055                                 & 0.063                               & 0.082                            & 0.348       & \ul{0.993}  & 0.864       & 0.802       & 0.858       & \ul{8.389}   & \ul{0.0404}  \\
      NRT                   & 0.672                           & 0.051                                 & 0.063                               & 0.075                            & \ul{0.384}  & \ul{0.993}  & \ul{0.894}  & 0.804       & \ul{0.867}  & 8.331        & 0.0402       \\
      PETER                 & 0.704                           & \ul{0.062}                            & 0.057                               & \ul{0.096}                       & 0.263       & 0.991       & 0.733       & \ul{0.819}  & 0.858       & 8.173        & 0.0239       \\
      PEPLER                & 0.661                           & 0.047                                 & 0.047                               & 0.087                            & 0.301       & 0.991       & 0.797       & 0.365       & 0.412       & 8.241        & 0.0061       \\
      ERRA                  & \ul{0.775}                      & 0.060                                 & \ul{0.064}                          & \tbf{0.097}                      & 0.277       & 0.917       & 0.859       & \tbf{0.825} & \tbf{0.868} & 8.316        & 0.0267       \\
      MAPLE                 & \tbf{0.807}                     & \tbf{0.185}                           & \tbf{0.108}                         & 0.087                            & \tbf{0.951} & \tbf{0.999} & \tbf{0.997} & 0.684       & 0.808       & \tbf{11.015} & \tbf{0.0699} \\
      \cdashline{0-11}[2pt/2pt]
      MAPLE-GT              & 0.684                           & 0.148                                 & 0.086                               & 0.167                            & 0.475       & 0.994       & 0.890       & 0.801       & 0.867       & 9.362        & 0.0506       \\ \hline
    \end{tabular}}
  \vspace{0.2cm}
  \resizebox{0.85\textwidth}{!}{
    \begin{tabular}{c|c|ccc|cccccc|c}
      \multicolumn{12}{c}{\tbf{Yelp23}}                                                                                                                                                                                                                                            \\ \hline
      \multicolumn{1}{c|}{} & \multicolumn{1}{c|}{Factuality} & \multicolumn{3}{c|}{Aspect-wise Exp.} & \multicolumn{6}{c|}{Text Diversity} & \multicolumn{1}{c}{Gen. Quality}                                                                                                     \\ 
      \tbf{Method}          & \tbf{iFMR}                      & \tbf{FCR}                             & \tbf{iFCR}                          & \tbf{GT-FMR}                     & \tbf{USR}   & \tbf{uUSR}  & \tbf{iUSR}  & \tbf{D-2}   & \tbf{D-3}   & \tbf{ENTR}   & \tbf{MAUVE}  \\ \hline
      Att2seq               & 0.618                           & 0.012                                 & 0.022                               & 0.230                            & 0.253       & \ul{0.987}  & \ul{0.977}  & \ul{0.878}  & 0.851       & 8.024        & \ul{0.0634}  \\
      NRT                   & 0.615                           & 0.009                                 & 0.021                               & 0.224                            & 0.216       & 0.975       & 0.956       & 0.876       & 0.850       & 7.657        & 0.0486       \\
      PETER                 & 0.678                           & 0.015                                 & 0.024                               & \ul{0.251}                       & 0.272       & 0.962       & 0.934       & 0.867       & 0.847       & 8.022        & 0.0604       \\
      PEPLER                & 0.690                           & 0.013                                 & 0.025                               & \tbf{0.261}                      & 0.303       & 0.950       & 0.911       & 0.809       & 0.799       & 8.311        & 0.0426       \\
      ERRA                  & \ul{0.775}                      & \ul{0.033}                            & \ul{0.03}                           & 0.181                            & \ul{0.388}  & 0.963       & 0.858       & \tbf{0.884} & \tbf{0.878} & \ul{8.992}   & 0.0267       \\
      MAPLE                 & \tbf{0.794}                     & \tbf{0.063}                           & \tbf{0.041}                         & 0.206                            & \tbf{0.855} & \tbf{0.999} & \tbf{0.997} & 0.816       & \ul{0.866}  & \tbf{11.485} & \tbf{0.2183} \\
      \cdashline{0-11}[2pt/2pt]
      MAPLE-GT              & 0.691                           & 0.039                                 & 0.029                               & 0.333                            & 0.620       & 0.998       & 0.994       & 0.892       & 0.886       & 10.316       & 0.1497       \\ \hline
    \end{tabular}}
  \caption{Automatic evaluation results on 10,000 user-item pairs sampled from the test sets. The "i"-prefix is "item-wise" for short, and the "u"-prefix is "user-wise". The best and second-best performances are bold-faced and underlined, respectively. Since MAPLE-GT employs ground-truth aspects, we do not put this variant into comparisons. Unless otherwise stated, MAPLE in the paper defaults to the Supervised@3 aspect-recommendation strategy.}
  \label{tab:main-results}
\end{table*}
\subsection{Automatic Evaluation Metrics}
\subsubsection{Diversity}
We measure the diversity of generated sentences using token, sentence, and corpus-level evaluations. For token-based metrics, we use the \tbf{Distinct-N} metrics by \citet{li2016diversity}.
To assess the severity of repetitive sentences generated by the model, we utilize the \tbf{Unique-Sentence Ratio (USR)} \citet{li2020nete} and the derived metrics, \tbf{user-wise USR} and \tbf{item-wise USR}\footnote{User-wise USR is calculated as the average USR ratio across a user's generated explanations, with the same approach applied for item-wise USR.}.
For the corpus-level, we employ the entropy-based metric ENTR by \citet{jhamtani2018entr}.

\subsubsection{Factuality}
For factuality evaluation, \citet{xie2023prag} introduces the \textit{Entailment Ratio}, which measures how many statements are supported by existing reviews of the same item. However, the generated content can synthesize aspects and opinions from multiple reviews, which the \textit{Entailment Ratio} metric may incorrectly classify as "unfactual." To address the above issue, we propose \tbf{Item-wise Feature Matching Ratio (iFMR)}. We treat the collection of extracted features\footnote{We use "features" interchangeably with "aspect terms" as \citet{li2020nete} refer to aspect terms as features in their proposed metrics.} in one restaurant's reviews as its "menu"; our rationale is that if the generated content contains any feature from the menu, it is "factual." Formally,
$$\text{iFMR} = \frac{1}{N} \sum_{u,i}^{N} \delta \left( \exists f\in F_i : f \in \hat{E}_{u,i} \right)$$
\( \hat{E}_{u,i} \) is the generated sentence for a user-item pair; \( F_i \) is the set of features associated with item \( i \) extracted from all training set reviews of that item; \( \delta(x) = 1 \) if \( x \) is true and \( \delta(x) = 0 \) otherwise.

\subsubsection{Precision}
To evaluate precision, which can also be interpreted as aspect-wise explainability, we adopt two metrics from \citet{li2020nete}: \tbf{Feature Coverage Ratio (FCR)} and \tbf{Ground-Truth Feature Matching Ratio (GT-FMR)}. FCR is computed as the number of distinct features contained in all the generated explanations, divided by the total number of the features in the training set\footnote{Since we evaluate only a test subset, we count the features associated with its items and use the total as the denominator.}. \tbf{Ground-Truth Feature Matching Ratio (GT-FMR)} measures whether a generated explanation contains the feature in the ground-truth text.\footnote{It was originally proposed as a Feature-Matching Ratio (FMR). To distinguish it from the factuality metric item-wise FMR, we prefix it with "GT".}
Additionally, it's important to consider the perspective of restaurant owners, who invariably prefer that a greater number of their restaurant’s advantages be highlighted and recommended to users in the explanation texts. Therefore, we craft \tbf{item-wise Feature Coverage Ratio (iFCR)} to provide insights into how well a model covers the relevant features for each item based on the training data.
For more implementation details and concrete formulas, refer to Appendix \ref{appenix:auto-metrics}.

\subsubsection{Generation Quality}

Unlike targeted generation tasks such as text summarization, explanation generation is a form of open-ended text generation. This nature makes the use of metrics like BLEU or ROUGE, which rely on exact token-by-token matches, less suitable.  To address this, we employ \tbf{MAUVE} \cite{Pillutla2021MAUVE} to quantify the gap between generated explanations and human texts, providing a statistical measure of generation quality.

\subsection{Baselines}
We compare our model against two groups of baseline methods.
The first group consists of classic end-to-end review generation models focused on NLG for explanations. We cover \tbf{Att2seq} \citep{Li2017att2seq}, \tbf{NRT} \citep{Li2017nrt}, \tbf{PETER} \citep{li2021peter}, \tbf{PEPLER} \citep{li2023pepler}, and \tbf{ERRA} \citep{ERRA2023explainable}. The second group is the retriever architecture in the retriever-reader framework, in which, to the best of our knowledge, only the personalized retriever in \tbf{PRAG} \citep{xie2023prag} is available. For more details, see Appendix \ref{appendix:baselines}.
\section{Results and Analysis}
\subsection{Quantitative Analysis on Explanations}
In our quantitative analysis, we compare MAPLE with baseline review-generation models that use item and user IDs but lack aspect input signals. Despite not multitasking on overall rating prediction, MAPLE significantly outperforms in factuality, feature coverage, and both sentence-level and corpus-level diversity. It shows more than a 10\% improvement in item-wise FMR, recommending accurate features in about 80\% of explanations. While striving for precision, such as choosing specific features like "salmon sushi" over general "food," MAPLE experiences a trade-off between FCR and FMR, especially notable in its lower GT-FMR score on Yelp23. Nonetheless, it excels in textual diversity on Yelp19, evidenced by high USR and ENTR scores, indicating minimal sentence repetition and high creativity. Besides, MAPLE also achieved high MAUVE scores, demonstrating a generation distribution closely aligned with human texts. For qualitative analysis and case studies, refer to Appendix \ref{appendix:case-study}.

\subsection{Ablation Study on Aspect Recommendation Strategies}
\begin{table*}[t]
  \small
  \centering
  \renewcommand\arraystretch{1.05}
  \resizebox{0.85\textwidth}{!}{
    \begin{tabular}{c|c|cc|ccc|c|c|c}
      \hline
      \multicolumn{10}{c}{\tbf{Yelp19}}                                                                                                                                                                                            \\ \hline
      \multicolumn{1}{c|}{} & \multicolumn{1}{c|}{Factuality} & \multicolumn{2}{c|}{Aspect-wise Exp.} & \multicolumn{3}{c|}{Text Diversity} & {Gen. Qty.} & {Text Sim.} & {Ranking}                                                \\ 
      \tbf{Method}          & \tbf{iFMR}                      & \tbf{iFCR}                            & \tbf{GT-FMR}                        & \tbf{USR}   & \tbf{D-2}   & \tbf{ENTR}   & \tbf{MAUVE}  & \tbf{BLEU-4} & \tbf{HR}    \\ \hline
      S@1                   & 0.728                           & 0.077                                 & 0.061                               & 0.654       & \ul{0.830}  & 10.580       & 0.0560       & \ul{0.297}   & 0.350       \\
      S@2                   & 0.789                           & 0.099                                 & 0.081                               & 0.840       & 0.768       & \ul{11.094}  & 0.0451       & 0.245        & 0.568       \\
      S@3                   & \ul{0.805}                      & \ul{0.108}                            & 0.087                               & \ul{0.944}  & 0.721       & 10.958       & \ul{0.0699}  & 0.238        & \ul{0.716}  \\
      S@4                   & \tbf{0.823}                     & \tbf{0.125}                           & \ul{0.096}                          & \tbf{0.971} & 0.666       & \tbf{11.497} & \tbf{0.0813} & 0.182        & \tbf{0.806} \\ 
      Heuristic@3           & 0.731                           & 0.094                                 & 0.076                               & 0.942       & 0.733       & 10.998       & 0.0504       & 0.230        & 0.419       \\
      \cdashline{0-9}[2pt/2pt]
      GT@1                  & 0.684                           & 0.086                                 & \tbf{0.167}                         & 0.728       & \tbf{0.833} & 8.041        & 0.0506       & \tbf{0.585}  & -           \\ \hline
      \multicolumn{10}{c}{\tbf{Yelp23}}                                                                                                                                                                                            \\ \hline
      S@1                   & 0.548                           & 0.023                                 & \ul{0.232}                          & 0.336       & \ul{0.877}  & 8.587        & 0.0548       & \ul{1.528}   & 0.710       \\
      S@2                   & 0.725                           & 0.035                                 & 0.200                               & 0.661       & 0.856       & \ul{11.487}  & \ul{0.2018}  & 0.952        & 0.863       \\
      S@3                   & \ul{0.794}                      & 0.041                                 & 0.207                               & \ul{0.855}  & 0.816       & 11.485       & \tbf{0.2183} & 0.947        & \ul{0.916}  \\
      S@4                   & \tbf{0.834}                     & \tbf{0.054}                           & 0.227                               & \tbf{0.921} & 0.734       & \tbf{12.182} & 0.1594       & 0.596        & \tbf{0.948} \\
      Heuristic@3           & 0.785                           & \ul{0.042}                            & 0.178                               & \ul{0.855}  & 0.816       & 11.441       & 0.1796       & 0.861        & 0.811       \\
      \cdashline{0-9}[2pt/2pt]
      GT@1                  & 0.691                           & 0.029                                 & \tbf{0.333}                         & 0.620       & \tbf{0.892} & 10.316       & 0.1497       & \tbf{2.616}  & -           \\ \hline
    \end{tabular}
  }
  \caption{Comparison of MAPLE with different aspect selection strategies. S@k denotes Supervised strategies with different values of $K$. HR is short for Hit Ratio. For Yelp23, there might be multiple ground-truth review segments; in this case, we report its BLEU-4 with  the multi-reference mechanism.}
  \label{tab:ablation-aspect-select}
\end{table*}
The trained aspect-recommendation component aims to select the best-fit aspect categories based on user-item IDs. While its performance is not directly evaluated, it plays a crucial role in determining the textual quality and topic of the generated content. We conduct an ablation study on the aspect-recommendation strategies with or without the component and display selected statistics in Table \ref{tab:ablation-aspect-select}.
\begin{enumerate}
  \item \tbf{Supervised@K (K=1,2,3)}: Inferencing the trained aspect-recommendation component to select $K$ aspects from a trimmed predicted aspect distribution (Appendix \ref{appendix:implementation}).
  \item \tbf{Heuristic@3}: Randomly sampling three categories from the intersection of item's and user's category histories.\footnote{Default to item history if the intersection size is smaller than three.}
  \item \tbf{GT@1}: Simulating the scenario where the user has either specified their interested aspect or the aspect-recommendation model is 100\% accurate by directly using the ground-truth aspect.\footnote{We do \textit{not} inference the trained aspect-recommendation component for Heuristic and GT.}
\end{enumerate}
\textbf{A proper mix of aspects enriches explanations.} Increasing the $K$ value in the model enhances the hit ratio, which boosts corpus-level diversity and feature coverage. Conversely, it slightly reduces token-level diversity, as evidenced by lower Distinct-2 scores, and significantly diminishes text-similarity scores. This reduction in Distinct-2 scores is attributed to the lengthening of sentences as $K$ increases. Given that the BLEU \citep{papineni2002bleu} is considered an indicator of textual relevance to the ground-truth reviews, we attempt to strike a balance between the diversity and textual relevance by monitoring BLEU-4 score. We employ the elbow method on BLEU-4 to identify the point where the slope sharply changes (when the component starts to produce overly generalized selection) and choose the point as our optimal $K$. We choose $K=3$ as the default value. Additionally, while heuristic selection strategies yield suboptimal results in aspect-wise explainability, the supervised approach is preferred due to its higher aspect ranking scores.\\
\begin{figure*}[t]
  \centering
  \includegraphics[width=0.8\textwidth]{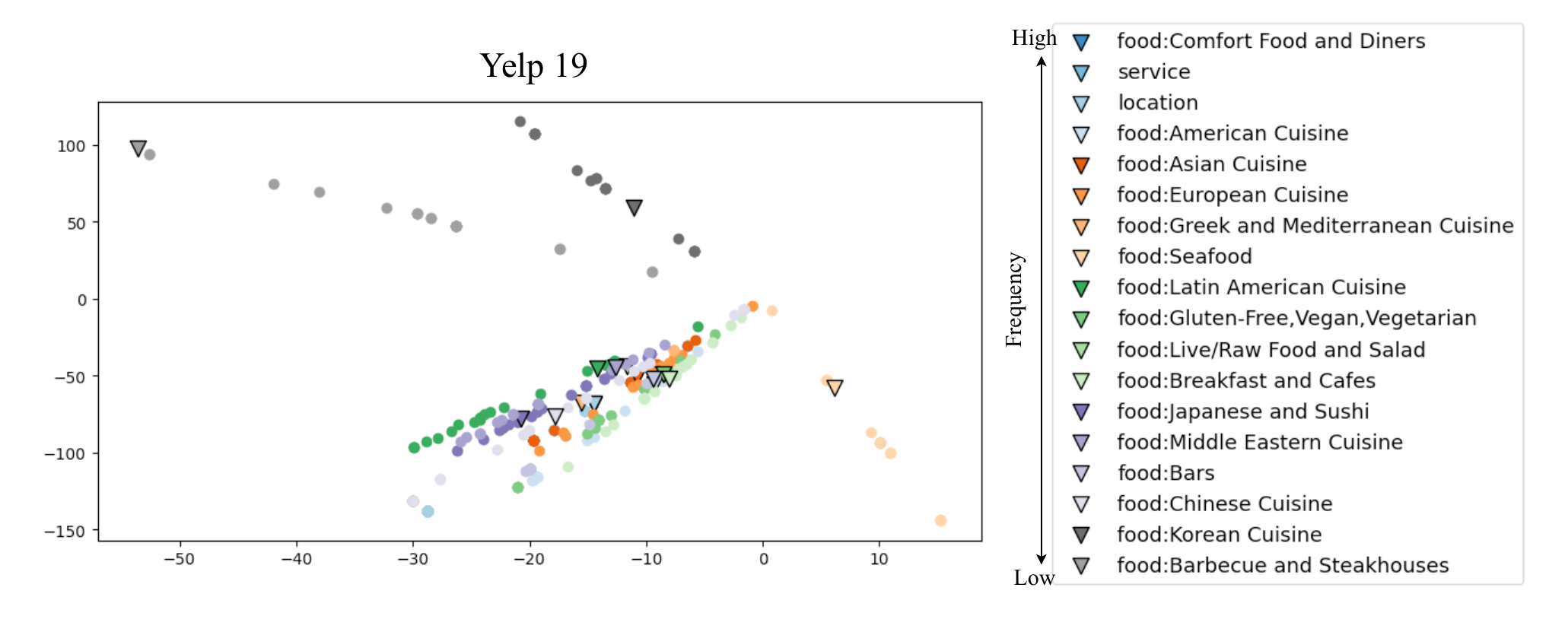}
  \caption{The t-SNE results on MAPLE trained on Yelp19 (fold 1) with S@3 strategy. The inverted triangle symbol indicates the aspect category prompt; the dots with the same color indicate the feature words belonging to the aspect category. We plot the 20 nearest feature words for each category.}
  \label{fig:t-sne}
\end{figure*}
\noindent\tbf{Ground-truth aspects lead to precise feature prediction and higher textual relevance.}
When provided with the ground-truth aspect category, the model's GT-FMR score significantly improves, demonstrating effective training of aspect prompts.
This improvement is attributed to the accurate aspect signal, which narrows the feature pool and thereby enhances feature prediction accuracy.
Furthermore, when conditioning on the ground-truth aspect, the BLEU-4 score tops all other strategies, showing that under the scenario where the user can provide the aspect category he/she is interested in, the explanation precision significantly improves. This underscores the value of precise user input in enhancing the relevance and accuracy of generated explanations.

\subsection{Explainability of Aspect Prompts}
Our hypothesis posits that MAPLE's aspect prompts tokens better retain and utilize less frequent features, which we explored using t-Distributed Stochastic Neighbor Embedding (t-SNE) \citep{Maaten2008tsne} to visualize the semantic clustering of feature words and aspect prompts. Figure \ref{fig:t-sne} shows linear clusters for the aspects and their feature words, confirming MAPLE's efficient learning of semantic associations. We further calculate the FCR scores by only considering the head 5 and the tail 5 aspect categories, denoted \tbf{head FCR} and \tbf{tail FCR}, and then present the statistics in Table \ref{tab:edge-features}. These results suggest that MAPLE's aspect prompt tokens significantly improve its ability to remember and utilize even the less frequent features, maintaining competitive performance even for less common aspect categories.
\begin{table}[H]
  \centering
  \small
  \renewcommand\arraystretch{1.00}
  \resizebox{0.35\textwidth}{!}{%
    \begin{tabular}{lccc}
      \toprule
      \multicolumn{4}{c}{\tbf{Yelp19}}                    \\
      \midrule
             & head FCR     & tail FCR     & FCR          \\
      \midrule
      PETER  & \ul{0.0635}  & \ul{0.0433}  & \ul{0.0616}  \\
      PEPLER & 0.0509       & 0.0237       & 0.0470       \\
      MAPLE  & \tbf{0.1993} & \tbf{0.1184} & \tbf{0.1846} \\
      \toprule
      \multicolumn{4}{c}{\tbf{Yelp23}}                    \\
      \midrule
             & head FCR     & tail FCR     & FCR          \\
      \midrule
      PETER  & \ul{0.0158}  & \ul{0.0099}  & \ul{0.0147}  \\
      PEPLER & 0.0141       & 0.0065       & 0.0131       \\
      MAPLE  & \tbf{0.0698} & \tbf{0.0353} & \tbf{0.0629} \\
      \bottomrule
    \end{tabular}
  }
  \caption{Head and tail FCR for MAPLE with sota baselines. FCR is attached for reference.}
  \label{tab:edge-features}
\end{table}
\subsection{MAPLE as a Discrete Retriever}
In this section, we examine the effectiveness of MAPLE as a discrete retriever within the retriever-reader framework, comparing it to the only known personalized retriever architecture from PRAG \citep{xie2023prag}. We evaluate both models from two perspectives: the \textit{latent} perspective and the \textit{aspect} perspective.

\noindent \tbf{Latent Perspective.}  We project MAPLE-generated explanations into the same latent space as PRAG's queries by encoding it with \texttt{all-mpnet-base-v2} to assess similarity against the ground-truth review, using Cosine Similarity and Mean Squared Error (MSE). Since PRAG's objective during training is to minimize the MSE, it performs very well in this context.

\noindent \tbf{Aspect Perspective.} 
We evaluate the model's ability to identify relevant aspects. MAPLE uses an aspect recommendation module targeting relevant losses, whereas for PRAG, we utilize its retrieved reviews from the user and item history, and collect the top-3 reviews' corresponding aspects as its prediction. Despite MAPLE's queries showing only 30\% similarity to ground-truth reviews, it more effectively identifies correct aspects, demonstrating stronger alignment with target goals.
Overall, PRAG excels in generating queries similar to ground-truth reviews, while MAPLE outperforms in accurately targeting relevant aspects.
\label{subsec:MAPLE as a discrete retriever}
\begin{table}[h!]
  \centering
  \renewcommand\arraystretch{1.1}
  \small
  \resizebox{0.4\textwidth}{!}{%
    \begin{tabular}{c|cc|cc}
      \hline
      \multirow{2}{*}{\textbf{Model}} & \multicolumn{2}{c|}{\tbf{Latent}} & \multicolumn{2}{c}{\tbf{Aspect}}                             \\
                                      & MSE                               & Cos Sim.                         & HR@3        & F1          \\
      \hline
      MAPLE                           & 0.002                             & 0.301                            & \tbf{0.716} & \tbf{0.369} \\
      PRAG                            & \tbf{0.001}                       & \tbf{0.439}                      & 0.389       & 0.228       \\
      \hline
    \end{tabular}
  }
  \caption{Comparison of Latent Metrics and Aspect Metrics for MAPLE and PRAG models}
  \label{tab:vs.prag}
\end{table}
\section{Conclusion}
We introduce the Multi-Aspect Prompt LEarner (MAPLE), a model that leverages user IDs and multi-aspect signals to generate detailed and controllable explanatory texts. Our primary technical achievements include the simple yet effective integration of aspect information into representation learning, aspect recommendation, and the learning of review sentences. We also prove that MAPLE can serve as a good discrete retriever in a retriever-reader explainable pipeline.
\section{Limitations}
Despite its high features and textual diversity, the MAPLE model presents several limitations. A notable challenge is the labeling of aspect categories. Although automated, this process still necessitates manual effort to define the aspect category inventory. The quality of these labels and their distribution across the dataset impact the training difficulty of the aspect recommendation component and subsequently the inference text quality and style. In cases where the label distribution is highly skewed, it might be necessary to optimize the aspect recommendation component separately. Additionally, our introduction of the item-wise Feature Matching Ratio marks a pioneering step towards enriching the aspect-wise factuality perspective of explainable recommendation model evaluations. While this metric adeptly identifies factual features in sentences, its capability to detect non-factual elements still needs future research and methodological advancement.

\section*{Acknowledgments}
We thank the Google Pixel Software Team for providing strong research support to this project.

\bibliography{custom}

\begin{thebibliography}{32}
\providecommand{\natexlab}[1]{#1}

\bibitem[{Althubyani et~al.(2024)Althubyani, Meng, Xie, Seung, Razzak, Sandoval, Kocaballi, Bamdad, and Naranjo}]{Pillutla2021MAUVE}
Mohammed Althubyani, Zhijin Meng, Shengyuan Xie, Cha Seung, Imran Razzak, Eduardo~Benitez Sandoval, Baki Kocaballi, Mahdi Bamdad, and Francisco~Cruz Naranjo. 2024.
\newblock \href {https://arxiv.org/abs/2412.04908} {Percy: A multimodal dataset and conversational system for personalized and emotionally aware human-robot interaction}.
\newblock \emph{Preprint}, arXiv:2412.04908.

\bibitem[{Cheng et~al.(2023)Cheng, Wang, Lu, Zhang, Zhou, Lu, and Liao}]{ERRA2023explainable}
Hao Cheng, Shuo Wang, Wensheng Lu, Wei Zhang, Mingyang Zhou, Kezhong Lu, and Hao Liao. 2023.
\newblock \href {https://doi.org/10.18653/v1/2023.acl-long.4} {Explainable recommendation with personalized review retrieval and aspect learning}.
\newblock In \emph{Proceedings of the 61st Annual Meeting of the Association for Computational Linguistics (Volume 1: Long Papers)}, pages 51--64, Toronto, Canada. Association for Computational Linguistics.

\bibitem[{Dong et~al.(2017)Dong, Huang, Wei, Lapata, Zhou, and Xu}]{Li2017att2seq}
Li~Dong, Shaohan Huang, Furu Wei, Mirella Lapata, Ming Zhou, and Ke~Xu. 2017.
\newblock \href {https://aclanthology.org/E17-1059} {Learning to generate product reviews from attributes}.
\newblock In \emph{Proceedings of the 15th Conference of the {E}uropean Chapter of the Association for Computational Linguistics: Volume 1, Long Papers}, pages 623--632, Valencia, Spain. Association for Computational Linguistics.

\bibitem[{He et~al.(2017)He, Lee, Ng, and Dahlmeier}]{He2017ABAE}
Ruidan He, Wee~Sun Lee, Hwee~Tou Ng, and Daniel Dahlmeier. 2017.
\newblock \href {https://doi.org/10.18653/v1/P17-1036} {An unsupervised neural attention model for aspect extraction}.
\newblock In \emph{Proceedings of the 55th Annual Meeting of the Association for Computational Linguistics (Volume 1: Long Papers)}, pages 388--397, Vancouver, Canada. Association for Computational Linguistics.

\bibitem[{Inc.(2023)}]{yelp2023dataset}
Yelp Inc. 2023.
\newblock \href {https://www.yelp.com/dataset} {Yelp open dataset}.
\newblock Accessed: 2023-10-15.

\bibitem[{Jhamtani et~al.(2018)Jhamtani, Gangal, Hovy, Neubig, and Berg-Kirkpatrick}]{jhamtani2018entr}
Harsh Jhamtani, Varun Gangal, Eduard Hovy, Graham Neubig, and Taylor Berg-Kirkpatrick. 2018.
\newblock \href {https://doi.org/10.18653/v1/P18-1154} {Learning to generate move-by-move commentary for chess games from large-scale social forum data}.
\newblock In \emph{Proceedings of the 56th Annual Meeting of the Association for Computational Linguistics (Volume 1: Long Papers)}, pages 1661--1671, Melbourne, Australia. Association for Computational Linguistics.

\bibitem[{Lewis et~al.(2020)Lewis, Liu, Goyal, Ghazvininejad, Mohamed, Levy, Stoyanov, and Zettlemoyer}]{Lewis2020bart}
Mike Lewis, Yinhan Liu, Naman Goyal, Marjan Ghazvininejad, Abdelrahman Mohamed, Omer Levy, Veselin Stoyanov, and Luke Zettlemoyer. 2020.
\newblock \href {https://doi.org/10.18653/v1/2020.acl-main.703} {{BART}: Denoising sequence-to-sequence pre-training for natural language generation, translation, and comprehension}.
\newblock In \emph{Proceedings of the 58th Annual Meeting of the Association for Computational Linguistics}, pages 7871--7880, Online. Association for Computational Linguistics.

\bibitem[{Li et~al.(2016)Li, Galley, Brockett, Gao, and Dolan}]{li2016diversity}
Jiwei Li, Michel Galley, Chris Brockett, Jianfeng Gao, and Bill Dolan. 2016.
\newblock \href {https://doi.org/10.18653/v1/N16-1014} {A diversity-promoting objective function for neural conversation models}.
\newblock In \emph{Proceedings of the 2016 Conference of the North {A}merican Chapter of the Association for Computational Linguistics: Human Language Technologies}, pages 110--119, San Diego, California. Association for Computational Linguistics.

\bibitem[{Li et~al.(2020)Li, Zhang, and Chen}]{li2020nete}
Lei Li, Yongfeng Zhang, and Li~Chen. 2020.
\newblock Generate neural template explanations for recommendation.
\newblock In \emph{Proceedings of the 29th ACM International Conference on Information \& Knowledge Management}, pages 755--764.

\bibitem[{Li et~al.(2021)Li, Zhang, and Chen}]{li2021peter}
Lei Li, Yongfeng Zhang, and Li~Chen. 2021.
\newblock \href {https://doi.org/10.18653/v1/2021.acl-long.383} {Personalized transformer for explainable recommendation}.
\newblock In \emph{Proceedings of the 59th Annual Meeting of the Association for Computational Linguistics and the 11th International Joint Conference on Natural Language Processing (Volume 1: Long Papers)}, pages 4947--4957, Online. Association for Computational Linguistics.

\bibitem[{Li et~al.(2023)Li, Zhang, and Chen}]{li2023pepler}
Lei Li, Yongfeng Zhang, and Li~Chen. 2023.
\newblock Personalized prompt learning for explainable recommendation.
\newblock \emph{ACM Transactions on Information Systems}, 41(4):1--26.

\bibitem[{Li et~al.(2017)Li, Wang, Ren, Bing, and Lam}]{Li2017nrt}
Piji Li, Zihao Wang, Zhaochun Ren, Lidong Bing, and Wai Lam. 2017.
\newblock \href {https://doi.org/10.1145/3077136.3080822} {Neural rating regression with abstractive tips generation for recommendation}.
\newblock In \emph{Proceedings of the 40th International ACM SIGIR Conference on Research and Development in Information Retrieval}, SIGIR '17, page 345–354, New York, NY, USA. Association for Computing Machinery.

\bibitem[{Loshchilov and Hutter(2019)}]{Ilya2019AdamW}
Ilya Loshchilov and Frank Hutter. 2019.
\newblock \href {https://openreview.net/forum?id=Bkg6RiCqY7} {Decoupled weight decay regularization}.
\newblock In \emph{International Conference on Learning Representations}.

\bibitem[{Lu et~al.(2011)Lu, Ott, Cardie, and Tsou}]{Lu2011MASA}
Bin Lu, Myle Ott, Claire Cardie, and Benjamin~K. Tsou. 2011.
\newblock \href {https://doi.org/10.1109/ICDMW.2011.125} {Multi-aspect sentiment analysis with topic models}.
\newblock In \emph{2011 IEEE 11th International Conference on Data Mining Workshops}, pages 81--88.

\bibitem[{Lu et~al.(2022)Lu, Liu, Dai, Xiao, Lin, Han, Sun, and Wu}]{lu-etal-2022-unified}
Yaojie Lu, Qing Liu, Dai Dai, Xinyan Xiao, Hongyu Lin, Xianpei Han, Le~Sun, and Hua Wu. 2022.
\newblock \href {https://aclanthology.org/2022.acl-long.395} {Unified structure generation for universal information extraction}.
\newblock In \emph{Proceedings of the 60th Annual Meeting of the Association for Computational Linguistics (Volume 1: Long Papers)}, pages 5755--5772, Dublin, Ireland. Association for Computational Linguistics.

\bibitem[{Maynez et~al.(2020)Maynez, Narayan, Bohnet, and McDonald}]{maynez2020faithfulness}
Joshua Maynez, Shashi Narayan, Bernd Bohnet, and Ryan McDonald. 2020.
\newblock \href {https://doi.org/10.18653/v1/2020.acl-main.173} {On faithfulness and factuality in abstractive summarization}.
\newblock In \emph{Proceedings of the 58th Annual Meeting of the Association for Computational Linguistics}, pages 1906--1919, Online. Association for Computational Linguistics.

\bibitem[{Ni and McAuley(2018)}]{Ni2018ExpansionNet}
Jianmo Ni and Julian McAuley. 2018.
\newblock \href {https://doi.org/10.18653/v1/P18-2112} {Personalized review generation by expanding phrases and attending on aspect-aware representations}.
\newblock In \emph{Proceedings of the 56th Annual Meeting of the Association for Computational Linguistics (Volume 2: Short Papers)}, pages 706--711, Melbourne, Australia. Association for Computational Linguistics.

\bibitem[{Papineni et~al.(2002)Papineni, Roukos, Ward, and Zhu}]{papineni2002bleu}
Kishore Papineni, Salim Roukos, Todd Ward, and Wei-Jing Zhu. 2002.
\newblock \href {https://doi.org/10.3115/1073083.1073135} {{B}leu: a method for automatic evaluation of machine translation}.
\newblock In \emph{Proceedings of the 40th Annual Meeting of the Association for Computational Linguistics}, pages 311--318, Philadelphia, Pennsylvania, USA. Association for Computational Linguistics.

\bibitem[{Pontiki et~al.(2016)Pontiki, Galanis, Papageorgiou, Androutsopoulos, Manandhar, Al-Smadi, Al-Ayyoub, Zhao, Qin, De~Clercq et~al.}]{pontiki2016semeval}
Maria Pontiki, Dimitrios Galanis, Haris Papageorgiou, Ion Androutsopoulos, Suresh Manandhar, Mohammad Al-Smadi, Mahmoud Al-Ayyoub, Yanyan Zhao, Bing Qin, Orph{\'e}e De~Clercq, et~al. 2016.
\newblock Semeval-2016 task 5: Aspect based sentiment analysis.
\newblock In \emph{10th International Workshop on Semantic Evaluation (SemEval 2016)}.

\bibitem[{Pontiki et~al.(2015)Pontiki, Galanis, Papageorgiou, Manandhar, and Androutsopoulos}]{pontiki2015semeval}
Maria Pontiki, Dimitrios Galanis, Harris Papageorgiou, Suresh Manandhar, and Ion Androutsopoulos. 2015.
\newblock Semeval-2015 task 12: Aspect based sentiment analysis.
\newblock In \emph{Proceedings of the 9th international workshop on semantic evaluation (SemEval 2015)}, pages 486--495.

\bibitem[{Pontiki et~al.(2014)Pontiki, Galanis, Pavlopoulos, Papageorgiou, Androutsopoulos, and Manandhar}]{pontiki2014semeval}
Maria Pontiki, Dimitris Galanis, John Pavlopoulos, Harris Papageorgiou, Ion Androutsopoulos, and Suresh Manandhar. 2014.
\newblock \href {https://doi.org/10.3115/v1/S14-2004} {{S}em{E}val-2014 task 4: Aspect based sentiment analysis}.
\newblock In \emph{Proceedings of the 8th International Workshop on Semantic Evaluation ({S}em{E}val 2014)}, pages 27--35, Dublin, Ireland. Association for Computational Linguistics.

\bibitem[{Raffel et~al.(2020)Raffel, Shazeer, Roberts, Lee, Narang, Matena, Zhou, Li, and Liu}]{raffel2020T5}
Colin Raffel, Noam Shazeer, Adam Roberts, Katherine Lee, Sharan Narang, Michael Matena, Yanqi Zhou, Wei Li, and Peter~J. Liu. 2020.
\newblock Exploring the limits of transfer learning with a unified text-to-text transformer.
\newblock \emph{J. Mach. Learn. Res.}, 21(1).

\bibitem[{Reimers and Gurevych(2019)}]{SentenceBERT}
Nils Reimers and Iryna Gurevych. 2019.
\newblock \href {https://arxiv.org/abs/1908.10084} {Sentence-bert: Sentence embeddings using siamese bert-networks}.
\newblock \emph{Preprint}, arXiv:1908.10084.

\bibitem[{Rorseth et~al.(2024)Rorseth, Godfrey, Golab, Srivastava, and Szlichta}]{10598017}
Joel Rorseth, Parke Godfrey, Lukasz Golab, Divesh Srivastava, and Jaroslaw Szlichta. 2024.
\newblock \href {https://doi.org/10.1109/ICDE60146.2024.00430} {{ RAGE Against the Machine: Retrieval-Augmented LLM Explanations }}.
\newblock In \emph{2024 IEEE 40th International Conference on Data Engineering (ICDE)}, pages 5469--5472, Los Alamitos, CA, USA. IEEE Computer Society.

\bibitem[{Sennrich et~al.(2016)Sennrich, Haddow, and Birch}]{Sennrich2016BPE}
Rico Sennrich, Barry Haddow, and Alexandra Birch. 2016.
\newblock \href {https://doi.org/10.18653/v1/P16-1162} {Neural machine translation of rare words with subword units}.
\newblock In \emph{Proceedings of the 54th Annual Meeting of the Association for Computational Linguistics (Volume 1: Long Papers)}, pages 1715--1725, Berlin, Germany. Association for Computational Linguistics.

\bibitem[{Sun et~al.(2021)Sun, Wu, Zhang, Su, and Wang}]{Sun2022UARM}
Peijie Sun, Le~Wu, Kun Zhang, Yu~Su, and Meng Wang. 2021.
\newblock \href {https://doi.org/10.1145/3483611} {An unsupervised aspect-aware recommendation model with explanation text generation}.
\newblock \emph{ACM Trans. Inf. Syst.}, 40(3).

\bibitem[{van~der Maaten and Hinton(2008)}]{Maaten2008tsne}
Laurens van~der Maaten and Geoffrey~E. Hinton. 2008.
\newblock \href {https://api.semanticscholar.org/CorpusID:5855042} {Visualizing data using t-sne}.
\newblock \emph{Journal of Machine Learning Research}, 9:2579--2605.

\bibitem[{Wu et~al.(2020)Wu, Huang, Liu, Wang, and Lin}]{Wu2020dbloss}
Tong Wu, Qingqiu Huang, Ziwei Liu, Yu~Wang, and Dahua Lin. 2020.
\newblock \href {https://doi.org/10.1007/978-3-030-58548-8_10} {Distribution-balanced loss for multi-label classification in long-tailed datasets}.
\newblock In \emph{Computer Vision – ECCV 2020: 16th European Conference, Glasgow, UK, August 23–28, 2020, Proceedings, Part IV}, page 162–178, Berlin, Heidelberg. Springer-Verlag.

\bibitem[{Xianghua et~al.(2013)Xianghua, Guo, Yanyan, and Zhiqiang}]{xianghua2013multi}
Fu~Xianghua, Liu Guo, Guo Yanyan, and Wang Zhiqiang. 2013.
\newblock Multi-aspect sentiment analysis for chinese online social reviews based on topic modeling and hownet lexicon.
\newblock \emph{Knowledge-Based Systems}, 37:186--195.

\bibitem[{Xie et~al.(2023)Xie, Singh, McAuley, and Majumder}]{xie2023prag}
Zhouhang Xie, Sameer Singh, Julian McAuley, and Bodhisattwa~Prasad Majumder. 2023.
\newblock Factual and informative review generation for explainable recommendation.
\newblock In \emph{Proceedings of the AAAI Conference on Artificial Intelligence}, pages 13816--13824.

\bibitem[{Zhang et~al.(2023)Zhang, Li, Deng, Bing, and Lam}]{Zhang2023ABSASurvey}
W.~Zhang, X.~Li, Y.~Deng, L.~Bing, and W.~Lam. 2023.
\newblock \href {https://doi.org/10.1109/TKDE.2022.3230975} {A survey on aspect-based sentiment analysis: Tasks, methods, and challenges}.
\newblock \emph{IEEE Transactions on Knowledge \&amp; Data Engineering}, 35(11):11019--11038.

\bibitem[{Zhang et~al.(2014)Zhang, Zhang, Zhang, Liu, and Ma}]{zhang2014sentires}
Yongfeng Zhang, Haochen Zhang, Min Zhang, Yiqun Liu, and Shaoping Ma. 2014.
\newblock \href {https://doi.org/10.1145/2600428.2609501} {Do users rate or review? boost phrase-level sentiment labeling with review-level sentiment classification}.
\newblock In \emph{Proceedings of the 37th international ACM SIGIR conference on Research \& development in information retrieval}, SIGIR '14, page 1027–1030, New York, NY, USA. Association for Computing Machinery.

\end{thebibliography}
\clearpage
\appendix
\section{Dataset Preparation}
\label{appendix:dataset-prep}
\subsection{Additional Details for Multi-Aspect Review Segmentation}
\label{appendix:dataset-models}
For the sentiment-analysis model, we use a T5-based fine-tuned information-extraction model \texttt{uie-large-en} \footnote{\url{https://huggingface.co/luyaojie/uie-large-en}}\citep{lu-etal-2022-unified} on SemEval datasets \cite{pontiki2014semeval,pontiki2015semeval,pontiki2016semeval}. For the zero-shot aspect-category classifier, we use the off-the-shelf \texttt{facebook/bart-large-mnli} released by Meta. We formulate the classification task as multiple Natural-Language Inference problems: for one aspect term, we ask the question \texttt{"\{aspect-term\}}. For each aspect category, we use the prompt \texttt{"This example is about \{category\}."}, and select the category with the highest entailment logit as the final aspect category for the given aspect term.\footnote{\url{https://huggingface.co/facebook/bart-large-mnli}} In this setup, we use T5 for precise span-level extraction, and BART for reliable category-level classification. We find this modular design to be more effective than relying on a single multi-task model.
\subsection{Additional Dataset Details}
\begin{figure*}[t]
  \centering
  \includegraphics[width=0.8\textwidth]{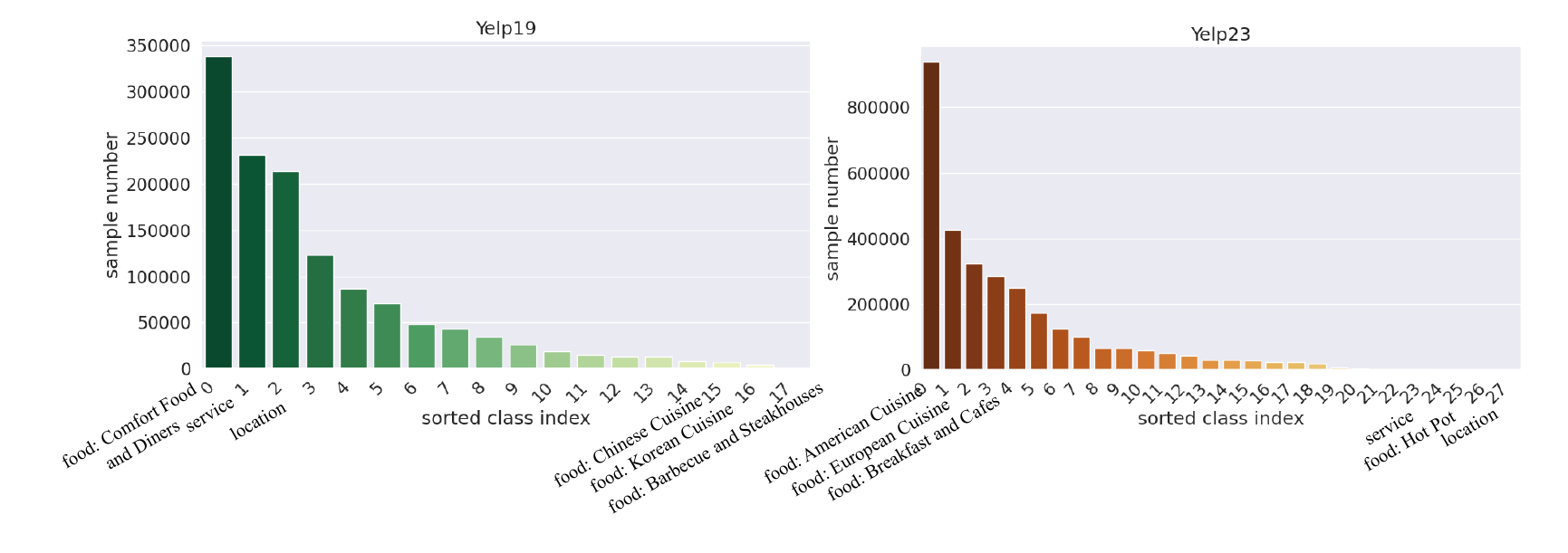}
  \caption{The long-tail dataset distribution of Yelp19 and Yelp23.}
  \label{fig:long-tail}
\end{figure*}
\label{appendix:dataset-details}

\begin{table}[H]
  \centering
  \resizebox{0.45\textwidth}{!}{%
    \begin{tabular}{l|c|c}
      \hline
                                & \tbf{Yelp19} & \tbf{Yelp23} \\\hline
      \# of users               & 27,147       & 35,152       \\\hline
      \# of items               & 20,266       & 24,199       \\\hline
      \# of reviews             & 1,293,247    & 1,339,433    \\\hline
      \# of features            & 204,117      & 279,636      \\\hline
      \# of segments            & 1,293,247    & 3,079,123    \\\hline
      \# of categories          & 18           & 28           \\\hline
      avg \# of reviews/user    & 47.64        & 38.10        \\\hline
      avg \# of reviews/item    & 63.81        & 55.35        \\\hline
      avg \# of segments/review & 1.0          & 2.298        \\\hline
    \end{tabular}
  }
  \caption{Dataset Statistics}
  \label{tab:dataset_statistics}
\end{table}

For Yelp23, we downloaded all the reviews from the online Yelp-hosted dataset website \footnote{\url{https://www.yelp.com/dataset}}. Since the raw data contains a large number of reviews, we follow the guidelines of \citet{li2020nete}. That is, we recursively prune users and items with fewer than 20 interactions, and split the datasets randomly into training, validation, and testing at the ratio of 8:1:1 5 times, while ensuring a warm-start scenario. Note that we split in terms of user-item pairs, i.e., all of the associated $(u, i,c_1), ..., (u, i, c_k)$ (k is the number of categories mined from the ground-truth review)
are guaranteed to appear in one of the training, validation, or testing stages.
For Yelp19, we utilize the data as processed by \citet{li2020nete}. Our sentiment analysis pipeline is applied to both Yelp19 and Yelp23. For Yelp23, we retain text spans illustrated in \ref{fig:sentiment-pipeline} as $E_{u,i,c}$. Since Yelp19 consists of only fragments of full reviews--which are not available to us--we treat the entire text associated with each user-item pair as $E_{u,i,c}$.

\subsection{Aspect Category Inventory}
We refer to the tags of items under Yelp website and SemEval workshop's aspect category inventory to define our aspect category inventory. We also consider the quality of the zero-shot labeling result and slightly alter the label list accordingly, which results in different aspect inventory for different datasets.
\label{appendix:dataset-aspectlist}
\begin{table*}[h!]
  \centering
  \begin{tabularx}{\textwidth}{|c|X|}
    \hline
    \tbf{Dataset} & \tbf{Categories}                                                \\
    \hline
    Yelp19        &
    American Cuisine, Asian Cuisine, Barbecue and Steakhouses, Bars, Breakfast and Cafes, Chinese Cuisine,
    Comfort Food and Diners, European Cuisine, "Gluten-Free, Vegan, Vegetarian", Greek and Mediterranean Cuisine,
    Japanese and Sushi, Korean Cuisine, Latin American Cuisine, Live/Raw Food and Salad, Middle Eastern Cuisine,
    Seafood, location, service                                                      \\
    \hline
    Yelp23        &
    African Cuisine, American Cuisine, Asian Cuisine, Barbecue and Steakhouses, Breakfast and Cafes, Burmese and Mongolian Cuisine,
    Chinese Cuisine, Comfort Food and Diners, European Cuisine, Food Court and Stands, Gastropubs and Modern European,
    "Gluten-Free, Vegan, Vegetarian", Greek and Mediterranean Cuisine, Halal and Kosher, Hot Pot, Japanese, and Sushi,
    Korean Cuisine, Latin American Cuisine, Live/Raw Food and Salad, Middle Eastern Cuisine, Seafood, South Asian Cuisine,
    Southeast Asian Cuisine, Tapas Bars, ambiance, location, miscellaneous; service \\
    \hline
  \end{tabularx}
  \caption{Aspect Inventories for Yelp19 \& 23 Datasets, with 18 and 28 categories respectively.}
  \label{tab:yelp_datasets}
\end{table*}




\section{Experiment Setup}
\subsection{Automatic Evaluation Metrics}
\label{appenix:auto-metrics}
\begin{itemize}
  \item \textbf{Feature Coverage Ratio (FCR)}: Measures the proportion of distinct features captured in the generated explanations compared to the total feature set.
        \[
          \text{FCR} = \frac{N_g}{|\mathcal{F}|}
        \]
        where \( \mathcal{F} \) represents the aggregated collection of features belonging to the sampled test items, and \( N_g \) is the number of distinct features shown in the generated explanations.

  \item \textbf{Item-wise Feature Coverage Ratio (iFCR)}: Calculates the average ratio of matched features to the total features for each item in the test set.
        \[
          \text{iFCR} = \frac{1}{M} \sum_{i \in \text{test\_set}} \frac{|f_{i}^{\text{matched}} \cap f_{i}^{\text{all}}|}{|f_{i}^{\text{all}}|}
        \]
        where \( M \) is the number of items in the test set, \( f_{i}^{\text{matched}} \) are the features matched in the generated explanations for item \( i \), and \( f_{i}^{\text{all}} \) are all the features associated with item \( i \) in the training set.

  \item \textbf{Ground-Truth Feature Matching Ratio (GT-FMR)}: Assesses whether the features in the generated explanations match those in the ground-truth text.
        \[
          \text{GT-FMR} = \frac{1}{N} \sum_{u,i}^{N} \delta \left( \exists f \in f_{u,i} : f \in \hat{E}_{u,i} \right)
        \]
        where \( \hat{E}_{u,i} \) is the generated sentence for a user-item pair, \( f_{u,i} \) is the set of features extracted from the ground truth review, and \( \delta(x) = 1 \) if \( x \) is true, \( \delta(x) = 0 \) otherwise.
\end{itemize}

For informativeness (iFMR) and some of the aspect-wise explainability metrics (iFCR, FCR), with an eye to encouraging more fine-grained aspect terms and penalizing the overly generic terms, we filter out the keywords that are too short (less than 4 characters), too general or noisy (see below list of \texttt{dummy\_words}).
\begin{verbatim}
dummy_words = ["and", "very", "the", "is",
"a", "an", "it", "this",
"that","of", "in", "that", "are",
"were", "was", "food"]
\end{verbatim}
\subsection{Additional Details for Baseline Models}
\label{appendix:baselines}
The first group consists of classic end-to-end review generation models focused on NLG for explanations.
\begin{itemize}
  \item \tbf{Att2Seq} \citep{Li2017att2seq}: An LSTM-based model.
  \item \tbf{NRT} \citep{Li2017nrt}: Another LSTM-based model designed for personalized review generation.
  \item \tbf{PETER} \citep{li2021peter}: An unpretrained transformer-based model that integrates user and item embeddings into the generation process.
  \item \tbf{PEPLER} \citep{li2023pepler}: A model leveraging a pre-trained GPT-2 with two-stage tuning for generating explanations. We specifically use the \textit{PEPLER-MF} variant, known for its superior text quality. Among these models, PEPLER’s architecture closely resembles ours, making it a crucial point of comparison.
  \item  \tbf{ERRA} \citep{ERRA2023explainable}: A model that combines both aspect-modeling and retrieval into the revised transformer architecture.
        While ERRA and MAPLE share similar core ideas, our model separates the retrieval process from itself and employs an LLM reader to comprehend the retrieved information from a broader perspective.
\end{itemize}
The second group of comparison methods involves models utilizing a retriever-reader framework.
\begin{itemize}
  \item \tbf{PRAG} \citep{xie2023prag}: A transformer-based model integrates user and item embeddings and review history. It employs a personalized attention mechanism to generate a latent query, which is optimized in an auto-encoder fashion against the embedding of the ground-truth review.
\end{itemize}
\section{Implementation for Experiments}
\subsection{Implementation Details}
\label{appendix:implementation}
MAPLE is trained on the training set with hyperparameters fine-tuned on the validation set. Evaluation is performed on 10,000 test-set pairs, averaged over 5 splits for MAPLE and review-generation models, while PRAG is tested only on the first split due to its long training time.
MAPLE is optimized using the AdamW optimizer \citep{Ilya2019AdamW}, with a learning rate of 0.001 and a batch size of 196. The maximum sequence length is limited to 20 Byte Pair Encoding (BPE) \citep{Sennrich2016BPE} tokens. The training process is divided into two stages as discussed in Section \ref{subsec:inference}: Stage 1 runs for up to 30 epochs with the tolerance times set to 5 (monitoring only the text loss); Stage 2 lasts for a maximum of 20 epochs with the tolerance times set to 2 (monitoring only the aspect-recommendation loss).
We use different training datassets for each stage. In the first stage, each user-item pair is associated with various aspect categories $c_i$ from the training set. In the second stage, focused on training the recommendation task, we trim the dataset so that each user-item pair appears only once.
The aspect regularization coefficient $\alpha$ is set to 0.01. The embedding dimension for user ID, item ID, and aspect is set to 768. For the aspect-recommendation component, we use an MLP with two hidden layers of 256 and 128 dimensions. Regarding the Distribution Balance loss, we set the negative-tolerance coefficient $\lambda$ to 1.0 and class bias $\nu_i$ to 0.05 for each class.

During inference, we apply greedy decoding for all models. For MAPLE, it sets $K$ to 3 as the number of aspects to be mixed as the guiding signal. For the personalized retriever of PRAG, we employ its item-marginalization variant and train with the user, item ID dimension of 768; the rest follow the default settings. For all the other baseline models, we adopt the official implementation as well as the default hyperparameter settings.
\subsection{Computational Resources and Cost Analysis}
All experiments are run on a single NVIDIA GeForce RTX 3090 (24 GB) GPU.

\textbf{Training and Inference Costs.} It takes about 14 and 7 hours to train MAPLE on Yelp23 and Yelp19, respectively. At inference, MAPLE performs aspect recommendation and then generates explanations conditioned on the predicted aspect vector. This adds \textit{only 0.02s latency per batch of 200 samples}, making it computationally comparable to baseline models (e.g., PETER, PEPLER, and ERRA) under the same backbone.

\textbf{+RAG Variant Resource Usage.} In case studies (Appendix~\ref{appendix:case-study}), the \textbf{MAPLE +RAG} variant performs retrieval by encoding all reviews using a sentence transformer and computing similarity scores over a review subset. This process requires approximately 60 GB of CPU RAM. For language generation, we use the \texttt{GPT-4} model via ChatGPT (accessed on June 10, 2024) with default settings and cleared history.
\begin{figure*}[ht]
  \centering
  \begin{verbatim}
  You are a restaurant recommendation explainer. Along with
  1. a user, 2. a restaurant, you are also given
  3. a model predicted user's personal query to this restaurant,
  which may not be a true statement,
  4. reviews about the restaurant and 5. reviews written by the users himself
  in the other restaurants.

  With the above information, you should pinpoint a feature within the "4. restaurant
  reviews", A feature can be a dish or an aspect (eg. service, location, etc.) of the
  restaurant. The feature MUST be mentioned in the restaurant reviews. If the personal
  query mentions a feature, you can use that. The user's reviews on the other restaurants
  do not hold true for the current restaurant. You should explain why the user might like
  or dislike the feature. The explanation should be short and concise within 50 words.
  Try to summarize the opinions if there are many discussing the same feature.
  Begin your explanation with
  "You may be interested in".

  User: {user} Restaurant: {item}
  Personal query: {personal_query}
  Restaurant reviews (where you find the feature to recommend the user):
  {item_reviews}
  User reviews (where you can refer to or identify user's preferences from):
  {user_reviews}
\end{verbatim}
  \caption{Prompt template for the reader model. In terms of the personal query, for MAPLE+RAG, we use the MAPLE-generated explanation; for PRAG+RAG, we use the translated and filtered keyword set.}
  \label{fig:prompt-template}
\end{figure*}
\section{Qualitative Case Studies}
\label{appendix:case-study}
\begin{figure*}[t]
  \includegraphics[width=\textwidth]{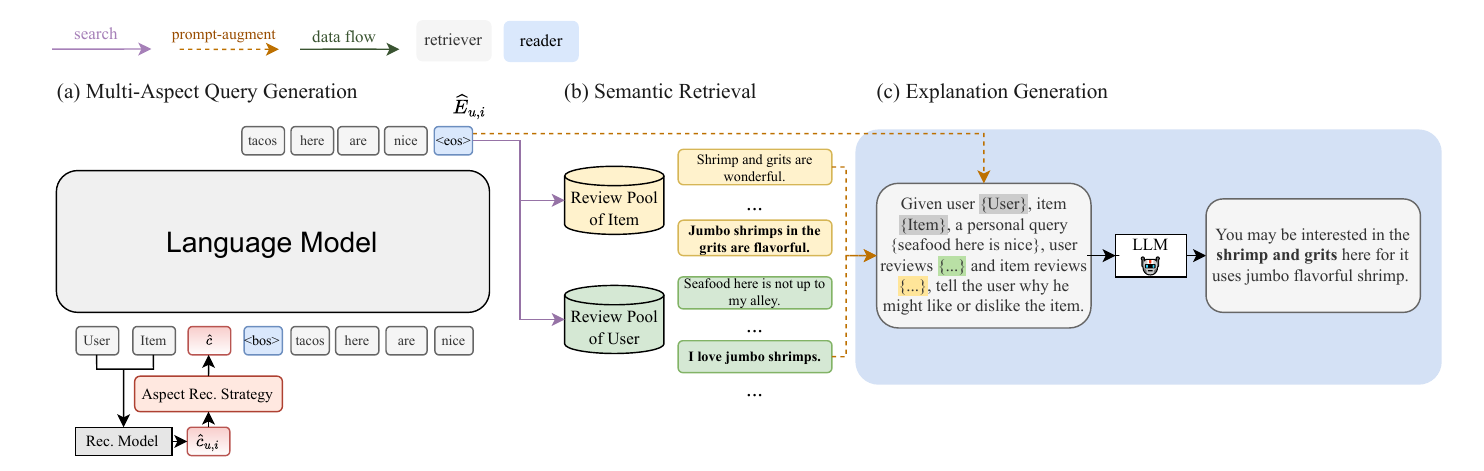}
  \caption{MAPLE and LLM in retriever-reader framework}
  \label{fig:maple+rag}
\end{figure*}

To address the issues of under-representation of user roles, we select several examples under the same item for the case study.
Since the personalized retriever in \citet{xie2023prag} produces a latent query for a user-item pair, we train the \textit{embedding interpreter} as \citet{xie2023prag} proposes to translate the query to a set of keywords: We input the latent query into the embedding interpreter and list the output keywords. In the implementation, it generates 5 keywords and then keeps only those appearing in the retrieved reviews, therefore, the displayed keywords may be fewer than five.
In the enhanced \tbf{+RAG} model, we treat the prefix model as the retriever and utilize \texttt{GPT-4} as the reader. We augment only MAPLE and PRAG with this setup.
For MAPLE, we encode its explanation with \texttt{all-mpnet-base-v2} \footnote{\url{https://huggingface.co/sentence-transformers/all-mpnet-base-v2}} and then treat them as queries to fetch the top-10 most similar reviews from the user and item history review pools, respectively. See \ref{fig:maple+rag} for illustration.
For PRAG personalized retriever, we use its proposed item-marginalization variant (i.e., subtract the mean of latent queries generated for the same item, and use it as the query embedding so as to highlight the /textit{distinctiveness} of each item) to retrieve pertinent reviews from the user and item pool.
\subsection{Case Study 1: Item Feature Precision}
It can be observed that in Table \ref{tab:case-study-1}, PRAG's retriever often produces queries focused narrowly on \tbf{service speed}, which lacks the breadth to capture the restaurant's primary attractions. Consequently, the \tbf{PRAG+RAG} mostly centers the discussions on service and wait times. Some more cases show that PRAG tends to prioritize service aspects.
Conversely, MAPLE excels in identifying detailed features such as ice cream \textit{flavors} (e.g., salted caramel, chocolate chips, strawberry), \textit{styles} (e.g., twist), and its \textit{complement} (e.g., the waffle cone), offering a variety of explanations tailored to different user preferences. In contrast, other review-generation models generally produce generic terms like "ice cream" or "toppings"; the best is probably PETER's "soft-served" in recommending user \tbf{xC-q\_yh0XwcjRLimkS3RNg}.
\subsection{Case Study 2: Personalization}
In exploring the personalization capabilities of recommendation systems, we focus on how effectively restaurant features are remembered and presented in user-specific contexts, i.e., how item IDs impact the generated explanations. A pertinent question arises: How about the impacts of user IDs? We assume an imaginary baseline approach, which simply samples reviews from an item’s review history. Could it possibly be as good as MAPLE or the other review-generation models? Or let's put it this way: what are the differences between it and the other review-generation models?
We observe that user signals primarily influence the \textit{style} and \textit{tone} of the generated sentences. For instance, in Table \ref{tab:case-study-french}, the user has a review history predominantly in French. Learning that the past reviews are primarily in French, the models generate sentences also in French. This level of personalization would be challenging to achieve with this random sampling baseline from the item’s review history. Could such a baseline possibly rival in terms of personalization of MAPLE or other user ID-incorporated review-generation models?
Moreover, MAPLE demonstrates an ability to retain and reflect personal details. For example, in Table \ref{tab:case-study-phoenix}, the user frequently mentions "Phoenix" in the review history, suggesting they might reside in the City of Phoenix. MAPLE captures this detail and incorporates it into the generated explanation, highlighting the user's local preferences. This ability to integrate personal context into recommendations underscores the model's strength in tailoring content to individual users.
These observations lead to the conclusion that the random baseline could not replicate the nuanced personalization achieved by models like MAPLE, which possess sophisticated mechanisms for personalizing generated sentences, adapting not only to users' linguistic preferences but also to subtle personal details, thereby enhancing the relevance and effectiveness of personalized recommendations.
\begin{table*}[h]
  \small
  \centering
  \begin{tabular}{l|p{11cm}}
    \hline
    \tbf{User}              & \tbf{Ground-Truth}                                                                                                                                                                                                                                                                                                                                                     \\\hline
    Hci2c0qo98CO-Pv-VmV7gg  & favorites are the bday cake one and smores                                                                                                                                                                                                                                                                                                                             \\
    \hline
    \tbf{Method}            & \tbf{Explanation}                                                                                                                                                                                                                                                                                                                                                      \\\hline
    \tbf{Att2seq}           & it 's a good \tbf{place} to get a quick bite to eat                                                                                                                                                                                                                                                                                                                    \\
    \tbf{NRT}               & the \tbf{ice cream} is good but the texture is a bit too sweet                                                                                                                                                                                                                                                                                                         \\        \tbf{PETER} & the \tbf{ice cream} is good \\
    \tbf{PEPLER}            & the \tbf{ice cream} is a bit on the sweet side                                                                                                                                                                                                                                                                                                                         \\\hline
    \tbf{PRAG}              & wait, parking, long                                                                                                                                                                                                                                                                                                                                                    \\

    \tbf{PRAG (+RAG)}       & You may be interested in the \tbf{quick service} despite the long lines at this restaurant. Reviews mention that even though the restaurant is small and the line can get long, the service is relatively quick.                                                                                                                                                       \\\hline
    \tbf{MAPLE}             & the \tbf{ice cream} is good but the \tbf{waffle cone} was a bit too soft for my liking                                                                                                                                                                                                                                                                                 \\
    \tbf{MAPLE (+RAG)}      & You may be interested in the \tbf{ice cream} at Sweet Jesus. Although reviews indicate the ice cream itself is average, the unique toppings and creative presentation could be appealing to you, especially given your past enjoyment of special flavors like black sesame and matcha in waffle cones.                                                                 \\
    \hline
    Ao-6FYE29-I8WwPg67806A  & the ice cream was nice and creamy and not dripping all over the place like most other soft-serve ice cream I've had                                                                                                                                                                                                                                                    \\
    \hline
    \tbf{Att2seq}           & it was a good \tbf{size}                                                                                                                                                                                                                                                                                                                                               \\
    \tbf{NRT}               & the \tbf{ice cream} was good                                                                                                                                                                                                                                                                                                                                           \\
    \tbf{PETER}             & the \tbf{ice cream} was good                                                                                                                                                                                                                                                                                                                                           \\
    \tbf{PEPLER}            & the \tbf{ice cream} was good but the \tbf{texture} was a little off                                                                                                                                                                                                                                                                                                    \\\hline
    \tbf{PRAG}              & line, fast, worth                                                                                                                                                                                                                                                                                                                                                      \\
    \tbf{PRAG (+RAG)}       & You may be interested in the \tbf{line management} and \tbf{atmosphere} at this restaurant. Reviews frequently mention long lines and crowded conditions, which may not be worth the wait for some.                                                                                                                                                                    \\\hline
    \tbf{MAPLE}             & i had the \tbf{salted caramel} and it was delicious                                                                                                                                                                                                                                                                                                                    \\
    \tbf{MAPLE (+RAG)}      & You may be interested in the \tbf{caramel offerings} at Sweet Jesus. You may enjoy the salted caramel, which aligns with the restaurant's praised caramel and coconut soft serve. Reviews consistently highlight their delicious caramel-focused desserts like the \tbf{sticky toffee pudding} and \tbf{caramel soft-serve}, suggesting these might match your tastes. \\
    \hline
    xC-q\_yh0XwcjRLimkS3RNg & raspberry puree and meringue crumble while the oh sweet baby jesus was caramel soft serve with peanut butter and caramel sauce                                                                                                                                                                                                                                         \\
    \hline
    \tbf{Att2seq}           & it 's a good \tbf{place} to get a quick bite to eat                                                                                                                                                                                                                                                                                                                    \\
    \tbf{NRT}               & the \tbf{ice cream} is good                                                                                                                                                                                                                                                                                                                                            \\
    \tbf{PETER}             & i 'm not a huge fan of the \tbf{soft serve ice cream}                                                                                                                                                                                                                                                                                                                  \\
    \tbf{PEPLER}            & the \tbf{ice cream} was good but the \tbf{texture} was a little off                                                                                                                                                                                                                                                                                                    \\\hline
    \tbf{PRAG}              & wasn't, line, huge, nice                                                                                                                                                                                                                                                                                                                                               \\
    \tbf{PRAG (+RAG)}       & You may be interested in the \tbf{quick service} and \tbf{small seating area} at Sweet Jesus. Although the restaurant is not large, reviews indicate that the line moves quickly and there is some outdoor seating, which might appeal to your preference for efficient service seen in your past reviews.                                                             \\\hline
    \tbf{MAPLE}             & i was craving something sweet so i ordered the \tbf{chocolate chip pizookie} which was a chocolate chip cookie                                                                                                                                                                                                                                                         \\
    \tbf{MAPLE (+RAG)}      & You may be interested in the \tbf{chocolate ice cream} at Sweet Jesus. Reviews suggest it's exceptionally chocolatey and comes with unique toppings like \tbf{marshmallows}. This seems to align with your love for sweet but well-balanced desserts, as noted in your previous reviews of desserts.                                                                   \\
    \hline
    flV6Cp3M2vHcqFBcCrGS1A  & i would definitely recommend getting the cone in a cup as it does tend to be messier than your regular ice cream                                                                                                                                                                                                                                                       \\
    \hline
    \tbf{Att2seq}           & the \tbf{ice cream} is good but \tbf{the ice cream} is a bit too sweet for my liking                                                                                                                                                                                                                                                                                   \\
    \tbf{NRT}               & the \tbf{ice cream} is a bit on the pricey side but the toppings are n't                                                                                                                                                                                                                                                                                               \\
    \tbf{PETER}             & the \tbf{ice cream} was good but the \tbf{ice cream} wasn't that great                                                                                                                                                                                                                                                                                                 \\
    \tbf{PEPLER}            & the \tbf{ice cream} was good but the \tbf{texture} was a little off                                                                                                                                                                                                                                                                                                    \\\hline
    \tbf{PRAG}              & plus, wait, large, pretty                                                                                                                                                                                                                                                                                                                                              \\
    \tbf{PRAG (+RAG)}       & You may be interested in the \tbf{efficiency of the line} at this restaurant. Despite the small size and often large crowds, reviews suggest that the line moves quickly and there's outdoor seating available. This might suit your preference for streamlined services, as seen in your past reviews.                                                                \\\hline
    \tbf{MAPLE}             & i've been here twice and i've tried the \tbf{red velvet and the chocolate twist}                                                                                                                                                                                                                                                                                       \\
    \tbf{MAPLE (+RAG)}      & You may be interested in the \tbf{"strawberry soft serve with caramel twist"} at Sweet Jesus. This unique flavor combination has been noted for its deliciousness and distinctiveness, possibly aligning with your adventurous taste in ice cream flavors.                                                                                                             \\
    \hline
  \end{tabular}
  \caption{Case studies on item id \tbf{N\_2yEZ41g9zDW\_gWArFiHw} (Sweet Jesus, an ice-cream shop) in Yelp19. There are a total of 35 reviews in the 10,000 test samples, 4 reviews are selected for demonstration. The recommended aspect terms mentioned in the explanations are boldfaced. }
  \label{tab:case-study-1}
\end{table*}
\clearpage
\begin{table*}[h]
  \small
  \centering
  \begin{tabular}{l|p{11cm}}
    \hline
    \tbf{User}             & \tbf{Ground-Truth}                                                                                                                                                                                                                                                                                                          \\\hline
    m-2OHy6aP26oefPIiiTocg & service is super quick but damn their food is good                                                                                                                                                                                                                                                                          \\
    \hline
    \tbf{Method}           & \tbf{Explanation}                                                                                                                                                                                                                                                                                                           \\\hline
    \tbf{Att2seq}          & le \tbf{service} est rapide et les serveurs sont                                                                                                                                                                                                                                                                            \\ 
    \tbf{NRT}              & le \tbf{service} est rapide et les plats sont                                                                                                                                                                                                                                                                               \\ 
    \tbf{PETER}            & le \tbf{service} est très agréable et le \tbf{service} est très agréable                                                                                                                                                                                                                                                    \\ 
    \tbf{PEPLER-MF}        & le \tbf{service} est the same .                                                                                                                                                                                                                                                                                             \\ 
    \hline
    \tbf{PRAG}             & delish, duck, bbq, pork, sweet                                                                                                                                                                                                                                                                                              \\ 
    \tbf{PRAG (+RAG)}      & You may be interested in trying the roasted BBQ pork at this restaurant. Reviews mention it as "sweet and well roasted" and "delicious," aligning with your query interests in BBQ and delicious dishes. This choice could offer you a comforting and flavorful dining experience.                                          \\ 
    \hline
    \tbf{MAPLE}            & le \tbf{menu} est original , vous souhaitez manger un peu difficileux et authentiquiri avec la qualit et la qualit et les banqu simplement \tbf{d'amoureux cocheureuse}                                                                                                                                                     \\ 
    \tbf{MAPLE (+RAG)}     & You may be interested in the \tbf{BBQ duck}, praised in the reviews for being sweet and well-roasted. It aligns with your query's desire for unique, high-quality meals. The restaurant's ability to surprise with unexpectedly good dishes despite their simple descriptions might also appeal to your adventurous palate. \\ %
    \hline
  \end{tabular}
  \caption{Case studies on item id \tbf{OllK5\_S-7svgSwbUfx1xYA} (King's Noodle Restaurant) in Yelp19. The recommended aspect terms mentioned in the explanations are boldfaced. }
  \label{tab:case-study-french}
\end{table*}
\begin{table*}[h]
  \small
  \centering
  \begin{tabular}{l|p{11cm}}
    \hline
    \tbf{User}             & \tbf{Ground-Truth}                                                                                                                                                                                                                                                                                                                                                                     \\\hline
    MG26Ips5arkOiFrvmmtcVg & the portions are small so you might want to order a side or appetizer with your entree                                                                                                                                                                                                                                                                                                 \\
    \hline
    \tbf{Method}           & \tbf{Explanation}                                                                                                                                                                                                                                                                                                                                                                      \\\hline
    \tbf{Att2seq}          & the \tbf{food} was great and the \tbf{service} was great                                                                                                                                                                                                                                                                                                                               \\ 
    \tbf{NRT}              & the \tbf{food }was good                                                                                                                                                                                                                                                                                                                                                                \\ 
    \tbf{PETER}            & the \tbf{food} is amazing                                                                                                                                                                                                                                                                                                                                                              \\ 
    \tbf{PEPLER-MF}        & the \tbf{food} is good                                                                                                                                                                                                                                                                                                                                                                 \\ 
    \hline
    \tbf{PRAG}             & spices, dish, amazing                                                                                                                                                                                                                                                                                                                                                                  \\ 
    \tbf{PRAG (+RAG)}      & You may be interested in the \tbf{food} at this restaurant, particularly for its \tbf{authentic Mexican flavors} and well-loved dishes such as \tbf{enchiladas} and \tbf{tacos}. Reviews consistently highlight the deliciousness and unique preparation of their meals, aligning with your appreciation for varied and tasty dishes noted in your past reviews.                       \\ 
    \hline
    \tbf{MAPLE}            & this is a great \tbf{place to relax} and enjoy the \tbf{beautiful \ul{Phoenix} sun}.                                                                                                                                                                                                                                                                                                   \\ 
                           & (Restaurant Reviews)                                                                                                                                                                                                                                                                                                                                                                   \\& - this is the place to go if you want some authentic Mexican food \\
                           & - a great local find                                                                                                                                                                                                                                                                                                                                                                   \\
                           & - the food was excellent                                                                                                                                                                                                                                                                                                                                                               \\
                           & - pretty good tacos                                                                                                                                                                                                                                                                                                                                                                    \\ & margaritas \\
                           & - and don't expect chips and salsa to arrive on your table when you sit down - it's not that kind of Mexican restaurant                                                                                                                                                                                                                                                                \\
                           & - it was fantastic; the enchiladas were nothing like traditional ones and were amazing.                                                                                                                                                                                                                                                                                                \\
                           & - unpretentious and positively delicious dishes prepared with love and taste like a practiced Mexican grandma labored in the kitchen to present you with an incredible meal                                                                                                                                                                                                            \\
                           & - what a great find                                                                                                                                                                                                                                                                                                                                                                    \\
                           & - what do you call two folded up pasty tortillas with unseasoned green sauce poured on top with an over abundance of crema and queso fresco                                                                                                                                                                                                                                            \\
                           & - we paid the nice lady and the prices were decent                                                                                                                                                                                                                                                                                                                                     \\&(User Reviews) \\&- this is going to be a popular spot in \ul{Phoenix} \\
                           & - the food is the best in \ul{Phoenix} and the service is always so friendly                                                                                                                                                                                                                                                                                                           \\
                           & - very happy we found this hidden gem in Tempe                                                                                                                                                                                                                                                                                                                                         \\
                           & - a perfect restaurant to bring the whole family                                                                                                                                                                                                                                                                                                                                       \\
                           & - such a wonderful neighborhood restaurant                                                                                                                                                                                                                                                                                                                                             \\
                           & - great place to try different dishes and they are all very good                                                                                                                                                                                                                                                                                                                       \\
                           & - a wonderful way to end your Sunday is with a campfire sundae!                                                                                                                                                                                                                                                                                                                        \\
                           & - the atmosphere however is a bit loud and their music makes it worse                                                                                                                                                                                                                                                                                                                  \\
                           & - make the trip and enjoy a wonderful breakfast with the family                                                                                                                                                                                                                                                                                                                        \\
                           & - the food was very good                                                                                                                                                                                                                                                                                                                                                               \\
    \hline
    \tbf{MAPLE (+RAG)}     & You may be interested in the \tbf{authentic Mexican food} at Escobar Mexican Kitchen. Reviewers consistently praise the quality and traditional flavors, describing dishes as "unpretentious and positively delicious," reminiscent of a seasoned Mexican grandma's cooking. This aligns well with your preference for places to relax and enjoy good food under \ul{the Phoenix sun}. \\ 
    \hline
  \end{tabular}
  \caption{Case studies on user id \tbf{MG26Ips5arkOiFrvmmtcVg} and item \tbf{Escobar Mexican Kitchen} in Yelp19. he recommended aspect terms mentioned in the explanations are boldfaced. The personalization details aligned in the explanation are underlined. }
  \label{tab:case-study-phoenix}
\end{table*}

\end{document}